\documentclass{article}
\usepackage[utf8]{inputenc}
\usepackage{graphicx} 
\usepackage[round,authoryear]{natbib}
\usepackage[a4paper, margin=1in]{geometry}
\usepackage[utf8]{inputenc}
\usepackage[T1]{fontenc}
\usepackage{microtype}
\usepackage[hyphens]{url} 
\usepackage{color}
\definecolor{darkblue}{rgb}{0.0, 0.0, 0.55}
\usepackage[colorlinks=true,linkcolor=darkblue,citecolor=darkblue,urlcolor=darkblue]{hyperref} 
\usepackage{comment}
\usepackage{booktabs} 
\usepackage{longtable}
\usepackage{pdflscape}  
\usepackage{geometry}   
\usepackage{amsmath}
\usepackage{listings}
\usepackage{array}
\usepackage{caption}
\usepackage{cleveref}
\usepackage{inconsolata} 

\definecolor{quotebg}{RGB}{245,245,250}
\definecolor{quoteborder}{RGB}{110,120,180}

\lstdefinestyle{feedbackprompt}{
  basicstyle=\footnotesize\ttfamily,
  backgroundcolor=\color{quotebg},
  frame=single,
  framesep=8pt,
  framerule=1pt,
  rulecolor=\color{quoteborder},
  breaklines=true,
  breakatwhitespace=false,
  captionpos=b,
  showstringspaces=false,
  numbers=none,
  xleftmargin=5pt,
  xrightmargin=5pt,
  aboveskip=10pt,
  belowskip=10pt
}

\usepackage{libertine} 

\title{Has the Creativity of Large-Language Models peaked? \\ --- an analysis of inter- and intra-LLM variability ---}

\author{
    Jennifer Haase\\
    Weizenbaum Institute and Humboldt University\\ Berlin, Germany\\
    \url{jennifer.haase@hu-berlin.de}
    \and
    Paul H. P. Hanel\\
    University of Essex, Colchester, UK\\
    \url{p.hanel@essex.ac.uk}
    \and
    Sebastian Pokutta\\
    TU Berlin and Zuse Institute Berlin \\ Berlin, Germany\\
    \url{pokutta@zib.de}
}

\date{\today}

\begin{document}

\maketitle



\begin{abstract}
Following the widespread adoption of ChatGPT in early 2023, numerous studies reported that large language models (LLMs) can match or even surpass human performance in creative tasks. However, it remains unclear whether LLMs have become more creative over time, and how consistent their creative output is. In this study, we evaluated 14 widely used LLMs---including GPT-4, Claude, Llama, Grok, Mistral, and DeepSeek---across two validated creativity assessments: the Divergent Association Task (DAT) and the Alternative Uses Task (AUT). Contrary to expectations, we found no evidence of increased creative performance over the past 18–24 months, with GPT-4 performing worse than in previous studies. For the more widely used AUT, all models performed on average better than the average human, with GPT-4o and o3-mini performing best. However, only 0.28\% of LLM-generated responses reached the top 10\% of human creativity benchmarks. Beyond inter-model differences, we document substantial intra-model variability: the same LLM, given the same prompt, can produce outputs ranging from below-average to original. This variability has important implications for both creativity research and practical applications. Ignoring such variability risks misjudging the creative potential of LLMs, either inflating or underestimating their capabilities. The choice of prompts affected LLMs differently. Our findings underscore the need for more nuanced evaluation frameworks and highlight the importance of model selection, prompt design, and repeated assessment when using Generative AI (GenAI) tools in creative contexts.\medskip

\textbf{Keywords:} Generative AI, benchmark testing, creativity, Large Language Models, LLMs
\end{abstract}

\section{Introduction}

Large Language Models (LLMs) have moved out of research labs and into our everyday lives. LLMs are often marketed as \textit{creative} (e.g., OpenAIs GPT 4.5, \citep{openai_introducing_2025} or Grok beta, \citep{xai_grok_2025}) because of their advanced abilities to generate text and respond to prompts. They allow users to brainstorm, draft content, and generate novel ideas with ease \citep{memmert_brainstorming_2024}. Consumers have responded eagerly, with recent surveys indicating that a majority of LLM users believe that these models enhance their creativity \citep{pandya_age_2024}. However, while LLMs can facilitate idea generation \citep{wan_it_2024, vaccaro_when_2024}, their widespread adoption raises important questions about their actual creative potential and the nature of their outputs \citep{runco_ai_2023}. Prior research suggests that LLM-generated ideas, while appearing individually creative, overall lead to homogenous outcomes across various domains, from creative writing and survey responses to research idea generation \citep{doshi_generative_2024, anderson_homogenization_2024, moon_homogenizing_2024}. For instance, stories written with ChatGPT assistance were more uniform than those generated independently by humans \citep{doshi_generative_2024}. Similarly, LLM-authored college essays contained fewer novel ideas than those written without LLM assistance \citep{moon_homogenizing_2024}. While these studies raise concerns about the creativity of LLMs, they typically focus on a single LLM, leaving unanswered the question of whether this homogeneity is unique to specific models or a broader phenomenon across different LLMs. Further, it is unclear whether LLMs have become more creative compared to 2023, when they became more widely known.

The present study aims to shed some light on these questions, in particular by addressing two key questions in this context: (1) Are current LLMs more creative than earlier versions and average human baselines, and which model performs best? (2) Do LLMs generate a diverse range of ideas within a session, or do their outputs converge toward homogeneous patterns? Put differently, are the answers of LLMs stable (i.e., homogeneous)? Previous research has neglected the output variability within the same LLM. This is important because greater variability (i.e., lower stability) within the responses of the same LLM can lead to either drastically over- or underestimating their creative capabilities. Recent research on feature space alignment in LLMs suggests that these models exhibit structural similarities, potentially leading to homogeneous outputs across different architectures \citep{wenger_were_2025, kleinberg_algorithmic_2021}. This would imply that regardless of the specific LLM used, users may experience a collective narrowing of creative expression due to shared underlying model biases \citep{huh_platonic_2024, lan_sparse_2025}. Because expectations and framing influence human creative performance, we also test whether prompting LLMs with the context of a ``creativity test'' affects their output quality, extending existing work on priming and task framing effects in LLMs \citep{gosling2024widely, renze2024effect, salinas2024butterfly}.

To explore these issues, we systematically evaluate a diverse set of LLMs using the Divergent Association Task (DAT) and the Alternative Uses Task (AUT), assessing both inter-model differences (creativity across models) and intra-model variance (creativity within repeated interactions with the same model). By using two tests as proxies to measure originality and semantic diversity, we examine which models best support creative tasks and whether they encourage or constrain the generation of diverse ideas.

As LLMs increasingly become integrated into the human-AI co-creative process, understanding the actual breadth and depth of their creative capacities is critical. Our findings have direct implications for model selection in practice, the design of collaborative tools, and the broader question of whether generative artificial intelligence (GenAI) systems meaningfully expand, or inadvertently narrow, the human creative landscape \citep{kleinberg_algorithmic_2021, doshi_generative_2024}. By revisiting and expanding previous research, we provide a comprehensive evaluation of LLM creativity, offering insights into their evolving role in human ideation and problem-solving.

\section{Creativity of Large Language Models}

Creativity has traditionally been regarded as a uniquely human trait: one that distinguishes us from machines and automation \citep{miller_artist_2019}. However, recent advances in GenAI have reignited debates about whether GenAI can exhibit creativity, particularly in fields such as literature, music, art, and problem-solving. While AI has already surpassed human capabilities in structured domains like Chess and Go \citep{gaessler_training_2023, krakowski_artificial_2023} and has been used in mathematics to solve open problems \citep{davies2021advancing, swirszcz2025advancing, MPSZ2024, MZKSP2025}, it remains uncertain whether it can achieve high levels of creativity or if it simply recombines existing knowledge in novel ways \citep{holford_future_2019, kirkpatrick_can_2023, white_opinion_2023}. Some argue that creativity remains one of the last strongholds of human superiority over AI \citep{holford_future_2019}, as it involves not only idea generation but also problem formulation, selection, and implementation \citep{botella_creative_2016, williams_mapping_2016}. Adding to the complexity of evaluating GenAI creativity is the well-documented sensitivity of LLMs to prompts. Small changes in phrasing or instruction framing can lead to substantial differences in output \citep{Mizrahi_state_2024, chang_survey_2024}, which complicates comparisons across studies and even within-task benchmarks. s (or LLMs) are instructed to generate ten words that are

Creativity---for humans---is defined as the ability to generate \textit{new} and \textit{useful} ideas \citep{runco_standard_2012, plucker_generalization_2004}. Further, it is conceptualized as an interaction between cognitive abilities, environmental factors, and social validation \citep{amabile_pursuit_2017}. High-level creativity, particularly in science and the arts, requires not only originality but also refinement, testing, and evaluation, as well as recognition so an idea is validated as a creative product \citep{benedek_motives_2020, kaufman_creative_2016, simonton_what_2013}. While human creativity involves free-associative thinking and problem formulation \citep{steele_looking_2018, botella_what_2018}, GenAI mimics these processes through probabilistic text generation and pattern recognition \citep{marcus_very_2022}. Although comparable in output as AI may produce things that are original and useful (cf. Section~\ref{C of LLM}), some argue that it does not ``create'' in the human sense: the difference lies not in what is produced but in how and why it is produced \citep{runco_ai_2023}.

Importantly, the discussion of AI's creative potential also addresses the ontological status of AI-generated creativity. Critics argue that GenAI, by virtue of relying on pre-existing data, is confined to ``incremental creativity'' and lacks the emotional depth and subjectivity that characterize human creative acts \citep{boden_computer_2009, cropley_creativity_2023, runco_ai_2023, white_opinion_2023}. According to this view, GenAI may simulate creativity convincingly but cannot embody the underlying processes of creative intention or self-expression.

We do not fully subscribe to such reductionist views. While it is true that LLMs are trained on existing knowledge, their ability to recombine, adapt, and contextualize information in novel ways demonstrates a form of inherent creativity. LLMs are designed to balance factual precision with creative expression, leveraging probabilistic language modeling, flexibility, and randomness to generate content that is perceived as original and inventive \citep{sinha_mathematical_2023, rafner_creativity_2023}. Empirical studies show that GenAI-generated outputs can sometimes match or even exceed human performance in tasks requiring originality and elaboration \citep{gilhooly_ai_2023, haase_artificial_2023}. Notably, in several domains, GenAI outputs are indistinguishable from human creations, successfully fooling experts in tasks ranging from scientific abstract writing \citep{else_abstracts_2023} to visual art production \citep{haase_art_2023}.

Although the philosophical debate about whether AI merely appears creative or truly \textit{is} creative remains unresolved \citep{runco_ai_2023, boden_computer_2009} and may not be practically relevant, our focus lies on \emph{empirical} outcomes and the \emph{measurable} creative potential of LLMs. Humans have created with the support of technology ever since the development of tools and technology \citep{haase_human-ai_2024}, and we aim to increase our understanding of how LLMs can be a creative support system by providing new and useful ideas to us as users. Thus, we briefly discuss how creativity is traditionally measured, what those measures reveal of LLMs' creative potential, and how this can be useful for the human co-creative process with LLMs in the next sections.

\subsection{Measuring Divergent Thinking}

Divergent thinking (DT) refers to the ability to generate multiple, varied, and novel ideas in response to open-ended problems. This ability is often perceived as a core cognitive process underlying creativity. Indeed, da \citet{da_costa_personal_2015}'s meta-analysis shows that DT demonstrates the strongest correlation with creativity-related constructs (\textit{\={r}}~= ~.27) when compared to other individual difference measures. However, this moderate effect size also illustrates a critical point: DT is far from synonymous with creativity. Rather, it should be seen as an indicator of creative potential, as one aspect of a broader and more complex cognitive and motivational landscape \citep{runco_four_2011}. One reason for the over-identification of DT with creativity lies in the dominant use of DT measures in creativity research. The most widely used measures---such as the Alternate Uses Test (AUT; \citealt{christensen_alternate_1960}) and the Torrance Tests of Creative Thinking (TTCT; \citealt{torrance_can_1972})---either directly assess divergent idea generation or embed association-based tasks as central components. As a result, much of what is empirically known about creativity relies on divergent idea generation. 

DT tasks are typically scored along several performance dimensions: \textit{fluency} (number of responses), \textit{flexibility} (variety of categories), \textit{originality} (statistical infrequency or uniqueness), and sometimes \textit{elaboration} (amount of detail). Among these, fluency is most often reported, although originality is arguably more aligned with creative value \citep{silvia_assessing_2008}. However, fluency scores tend to correlate strongly with originality, suggesting that originality may, at least in part, go along with response quantity in addition to quality. 

One standard way to measure DT is via the Alternate Use Task (AUT; \citealp{christensen_alternate_1960}) that asks participants to generate alternative uses for common objects. By contrast, the Divergent Association Task (DAT; \citealt{olson_naming_2021}), is a more recent measure that quantifies the semantic distance between ideas. Participants (or LLMs) are instructed to generate ten words that are as semantically different from each other as possible. Responses are then scored based on pairwise semantic distances obtained from a pre-trained embedding model, offering an index of creative divergence that is scalable and language-based. 

However, despite their widespread use, DT measures do not fully capture the complexity of creative thinking \citep{reiter-palmon_scoring_2019, runco_divergent_2012}. Creativity often requires the generation of ideas and the selection, refinement, and contextual evaluation of those ideas \citep{cromwell_creative_2022}. For example, \textit{convergent thinking} involves narrowing down multiple possibilities to identify the most effective solution and plays a key role in evaluating and implementing creative ideas \citep{cropley_praise_2006}. \textit{Emergent thinking}, on the other hand, involves exploring potential problem spaces for existing solutions, such as experimenting with new technologies to discover their applications \citep{cromwell_discovering_2023}. Recent work has also emphasized the importance of metacognitive strategies such as asking more complex questions, which can foster deeper and more creative exploration \citep{raz_role_2023}.

In sum, while DT remains a valuable proxy for \textit{creative potential}, particularly in controlled experimental contexts, it represents only one dimension of creative cognition. Its explanatory power is enhanced when situated within a broader framework that includes convergent, emergent, and metacognitive thinking. Accordingly, we use DT tasks in this study not as exhaustive indicators of creativity but as focused tools to probe one central aspect of the creative process.

\subsection{Empirical Analyses of LLMs' Creativity} \label{C of LLM}

Recent research suggests that GenAI, and LLMs in particular, are capable of producing outputs that meet established criteria for creativity---namely, \emph{originality} and \emph{usefulness} (e.g., \citealp{haase_artificial_2023, guzik_originality_2023}). In fields such as literature, GenAI-generated texts frequently match or even surpass human writing in fluency and coherence \citep{gomez-rodriguez_confederacy_2023}. Comparable evidence is emerging in other creative domains, including music \citep{civit_systematic_2022} and visual art \citep{dipaola_using_2016}. To empirically evaluate the creative potential of LLMs, researchers have increasingly turned to standardized psychological creativity assessments. These include divergent thinking tests such as the AUT, the DAT, and the TTCT (cf. Table~\ref{tab:creativity_overview} in the Appendix). However, it is important to note that LLM performance on such tasks is highly sensitive to prompt phrasing and format. Even subtle variations in instructions can lead to substantial differences in output quality and style, as LLMs rely on learned probabilistic patterns rather than intrinsic task comprehension \citep{chang_survey_2024}. In these tasks, GenAI systems such as GPT-3.5 and GPT-4 often perform at or above average human levels, particularly in dimensions like \emph{fluency} and \emph{elaboration} \citep{guzik_originality_2023, haase_artificial_2023}. Some studies even report that GPT-4 scores in the top 1\% in \emph{originality} and \emph{fluency} on the TTCT \citep{soroush_creativity_2025}. However, while LLMs frequently exceed average performance, they typically fall short of the originality levels exhibited by highly creative individuals or expert human creators \citep{koivisto_best_2023, haase_artificial_2023}. Further, although such findings are striking, they should be interpreted with caution, as model performance may reflect specific tuning to familiar prompt formats rather than generalized creative competence.

A parallel trend in this research area is the development of automated scoring systems that leverage AI and embedding-based metrics to evaluate creativity (e.g., \citealt{hadas_assessing_2025, organisciak_open_2025}). For example, the DAT relies on measuring semantic distance between generated words, and recent extensions include LLM-based scoring architectures \citep{haase2025synthetic}. Similarly, systems such as the Open-Ended Creativity Scoring AI (OpenScoring; \citealp{organisciak_beyond_2023}) apply transformer-based models to estimate creativity in free-form outputs, offering scalable alternatives to traditional human coding \citep{soroush_creativity_2025}. These approaches increase consistency and efficiency in the evaluation of creative output.

Taken together, the current state of evidence suggests that LLMs exhibit a level of creative potential that is both measurable and, in specific contexts, comparable to human creative output. While conceptual debates about the nature of creativity remain important, empirical evaluation provides a grounded basis for understanding the evolving role of GenAI in the overall creative process.

\subsection{Human-AI Co-Creativity}

Rather than replacing human creativity, GenAI has been shown to augment and enhance creative processes \citep{verganti_innovation_2020, haase_human-ai_2024, memmert_brainstorming_2024}. Collaborative work with tools like ChatGPT improves users' creative self-efficacy and has been associated with increased originality and elaboration in output \citep{urban_can_2023}. Human-AI partnerships are becoming increasingly common across creative domains such as screenwriting, music composition, and scientific problem-solving \citep{wan_it_2024, serbanescu_human-ai_2023}. In the arts, GenAI adoption correlates with a significant rise in productivity—artists using such tools report producing 50\% more artworks, alongside higher engagement rates from audiences \citep{zhou_generative_2024}. LLMs contribute to this development by offering ``beyond-human'' information processing capacities. Their ability to combine vast knowledge with stylistic variation enables outputs that are perceived as original and inventive \citep{sinha_mathematical_2023, soroush_creativity_2025}. When effectively used, LLMs support both fluency and originality in human output, allowing individuals to move more fluidly through ideation and elaboration phases \citep{doshi_generative_2024, heyman_supermind_2024}. They can also contribute in ways that exceed human capacity for complexity, suggesting non-intuitive solutions---as evidenced in research examples like plane coloring in mathematics \citep{MPSZ2024}, protein structures \citep{varadi_impact_2023} and autonomous driving \citep{Atakishiyev_explainable_2024}.

However, these benefits are accompanied by certain limitations. Longitudinal research indicates that the output of GenAI systems tends toward aesthetic homogeneity, with content appearing increasingly similar across instances and users \citep{zhou_generative_2024}. Moreover, AI-generated content often lacks the element of human surprise and subjectivity, offering elaborated yet convergent ideas that are well-structured but less likely to deviate from known paths \citep{song_interactions_2025, doshi_generative_2024}. In this regard, GenAI systems tend to reinforce structured problem-solving rather than facilitating radical or conceptual creativity. They excel at combinatorial creativity---that is, remixing and synthesizing existing knowledge---but still struggle with producing conceptual leaps or paradigm-shifting insights \citep{soroush_creativity_2025, orru_human-like_2023}. Accordingly, human input remains indispensable, not only for refining and evaluating the ideas produced but also for grounding them in contextually meaningful ways \citep{lazar_love_2022, runco_ai_2023}.

The most promising outcomes arise when humans and AI engage in mutual exploration. In such settings, the human creator defines the problem space and direction, the AI contributes associative or alternative paths, and both iterate toward refined and context-sensitive creative solutions \citep{haase_human-ai_2024}. This dynamic reflects well-established principles of team-based human creativity \citep{paulus_chapter_2012}, and its effectiveness is enhanced when interaction with AI is supported by thoughtful design, including mechanisms that preserve human control, ensure transparency, and encourage flexibility \citep{shneiderman_bridging_2020}. Furthermore, GenAI tools help democratize creativity by lowering the entry barriers for non-experts. They provide accessible means for ideation, scaffolding, and iterative refinement, even for individuals with limited prior experience \citep{rafner_creativity_2023}. This empowerment enables broader participation in creative domains, from education to entrepreneurship \citep{eapen_how_2023, cambon_early_2023}. Over time, such collaborative processes can even foster the long-term development of human creative capabilities. Just as professional Go players improved their strategic thinking through interactions with AlphaGo \citep{metz_sadness_2016}, users working iteratively with GenAI tools can develop more refined approaches to problem framing, idea evaluation, and creative synthesis.

In summary, GenAI has substantial potential to enhance human creativity when used as a co-creative collaborator rather than a substitute for human cognition. The key is to design and apply these systems in ways that augment, rather than constrain, human agency, enabling synergies that expand the boundaries of what is creatively possible. However, a basic prerequisite for this collaboration is the creative potential of LLMs, which we will test below.

\section{Experimental Setup}

To evaluate the creative potential of contemporary LLMs, we systematically compared multiple models on two standardized DT tasks --the DAT and the AUT-- assessing both inter-model performance and intra-model output variability. In addition, we tested two distinct prompt formulations for each task to examine whether prompt phrasing contributes to performance differences across models. While this is not a prompting-focused study as many more prompt variants could (and arguably should) be explored to fully capture systematic effects, we acknowledge that prompt design is a non-trivial factor influencing LLM output, particularly in open-ended generative tasks. As such, we include basic prompt variations to assess the robustness of model responses. The following section details the models evaluated, the software environment, and the methodology used for administering and scoring the tasks.

All experimental data were stored in JSON format. Each record contained metadata (e.g., timestamps, model identifiers, prompt variants), full results per model-object combination, and trial-level scores. The data and R-code to reproduce our analyses are publicly accessible via the Open Science Framework (OSF)\footnote{\url{https://osf.io/e62fx/?view_only=cd2f9763f84843c385ca760702eeae12}}. All models were accessed using their default parameter settings through their API via LiteLLM between 25 to 28 February 2025 for the DAT tests and 1 to 3 March 2025 for the AUT tests. Each trial was run in an isolated prompt session to avoid carryover effects or memory accumulation. To ensure reproducibility, all API calls were logged, and each model's version and associated metadata were recorded. 

\subsection{Models and Libraries}

As outlined, we used a broad range of widely used models in our experiments (summarized in Table \ref{tab:LLMs}), as research has shown that different models exhibit distinct tendencies in ``behavior'' even when prompted identically \citep{zhang_exploring_2024}.

\begin{table}[h]
\centering
\begin{tabular}{lll}
\toprule
\textbf{Model Name} & \textbf{Provider} & \textbf{Description} \\
\midrule
Claude-3.5-sonnet & Anthropic & Claude 3.5 Sonnet (20241022) \\
Claude-3.7-sonnet & Anthropic & Claude 3.7 Sonnet (20250219) \\
DeepSeek-R1-70B & DeepSeek & DeepSeek R1 70B\\
DeepSeek-R1-Distill-Qwen-7B & DeepSeek & DeepSeek R1 Distill Qwen 7B GGUF  \\
Gemini-pro & Google & Google's Gemini Pro model \\
GPT-4.5-preview & OpenAI & Preview version of GPT-4.5 \\
GPT-4o & OpenAI & GPT-4o, OpenAI's currently most capable multimodal model \\
Grok-2-latest & xAI & Latest version of Grok-2 \\
Grok-beta & xAI & xAI Beta channel model \\
Llama-3.3-70B-Instruct & Meta & Llama 3.3 70B Instruct GGUF \\
Mistral-Nemo-Instruct-2407 & Mistral & Mistral Nemo Instruct 2407 GGUF Q8\_0  \\
o3-mini & OpenAI & Smaller, reasoning model from O3 series \\
Phi-4 & Microsoft & Phi-4 GGUF  \\
Qwen-2.5-7B-Instruct-1M & Qwen & Qwen2.5 7B Instruct 1M GGUF \\
\bottomrule
\end{tabular}
\caption{\label{tab:LLMs}Large Language Models used in the experiments}
\end{table}

Our experimental setup was implemented in Python 3.13, using the following libraries:
\begin{itemize}
    \item \textbf{LiteLLM}: Version 1.x - Used for unified access to various LLM providers.
    \item \textbf{Pandas}: Version 2.x - Utilized for data manipulation and analysis.
    \item \textbf{Requests}: Employed for API calls and handling HTTP responses.
    \item \textbf{BeautifulSoup}: Used for parsing HTML responses where necessary.
\end{itemize}


\subsection{Divergent Association Task}
\subsubsection{Default Prompt Templates}
\label{sec:dat-prompts}

We used two different prompt templates for the DAT experiments to standardize input conditions across models. These prompts were designed to maximize the likelihood of generating highly semantically diverse word sets while maintaining clarity and conciseness. The difference between both prompts is whether the test situation of the DAT was disclosed (being-aware) or not (not-aware), with the expectation that, similar to humans, awareness of a test situation and priming to be creative influences the creative performance \citep{sassenberg_dont_2005, acar_what_2020}.

\paragraph{Prompt: Aware of being DAT tested}

We used the following prompt:

\begin{lstlisting}[style=feedbackprompt]
Generate 10 words for the Divergent Association Task (DAT). The goal is to 
generate words that are as different from each other as possible in all 
meanings and uses.

Rules:
Only single words in English
Only nouns (things, objects, concepts)
No proper nouns (no specific people or places)
No specialized vocabulary (no technical terms)
Words should be as semantically different from each other as possible
Please provide exactly 10 words, one per line, with no additional text or explanation.
\end{lstlisting}

\paragraph{Prompt: Not aware of being DAT tested}

We used the following prompt:

\begin{lstlisting}[style=feedbackprompt]
Generate 10 words. The goal is to generate words that are as different 
from each other as possible in all meanings and uses.

Rules:
Only single words in English
Only nouns (things, objects, concepts)
No proper nouns (no specific people or places)
No specialized vocabulary (no technical terms)
Words should be as semantically different from each other as possible
Please provide exactly 10 words, one per line, with no additional text or explanation.
\end{lstlisting}

\subsubsection{Scoring and Analysis}

The evaluation process involved submitting each model's generated words to the official DAT website recording the assigned scores, percentiles, and semantic distance matrices. The experimental process followed a structured workflow. First, the experimental setup was established by selecting the models, determining the number of trials, and defining the evaluation method. During the generation phase, each model produced 10 words based on the standardized prompt template. These words were then submitted to the DAT website for scoring, and the resulting scores, percentiles, and word matrices were recorded. Each model ran the test 100 times, not being aware of formerly done tests.

DAT scoring is based on semantic distance, resulting in a score that is the average semantic distance between all pairs of words (higher is better). Further, we baseline all scores with human-generated scores for better comparison. We use a large sample of humans with $n$ = 8,907 from \cite{olson_naming_2021}. We focus, unless otherwise stated, on the percentile rank, as it allows us to directly compare the responses of the LLMs with human responses (e.g., if percentile >~50, the LLM would be better than the average human). 

Following data collection, results were analyzed to assess model performance across the score metrics. The analysis included calculating and visualizing the average scores per model, examining score distributions, and constructing word matrices to illustrate semantic distances between generated words. The norming based on human performance allows for comparing how LLM outputs align with human divergent thinking performance. 

\subsection{Alternate Use Task}
LLMs are evaluated using the AUT to assess their ability to produce original and useful ideas. For scoring and evaluation, we used the OpenScoring API\footnote{\url{https://openscoring.du.edu/llm}} with default parameters. All documentation was referenced from the official API guide\footnote{\url{https://openscoring.du.edu/docs}}.

\subsubsection{Default Prompt Templates}
\label{sec:aut-prompts}

Two distinct prompt templates were used to assess whether the framing of the task would influence the creative output of the LLMs. The first prompt emphasized practicality and feasibility, while the second emphasized creative and unconventional thinking. Both prompts instructed the model to avoid duplications and to generate a specified number --100-- of responses:

\paragraph{Prompt: Practical and Feasible}
We used the following prompt:

\begin{lstlisting}[style=feedbackprompt]
Generate creative alternative uses for the common object: {prompt}.

Rules:
1. Each use should be different from the object's intended purpose
2. Uses should be practical and feasible
3. Avoid duplicating ideas
4. Provide brief but clear descriptions
5. Be creative and think outside the box

Please provide {num_responses} different uses, one per line, 
with no additional text or explanation.
\end{lstlisting}

\paragraph{Prompt: Creative and Unconventional}
We used the following prompt:

\begin{lstlisting}[style=feedbackprompt]
Generate creative alternative uses for the common object: {prompt}.

Rules:
1. Each use should be different from the object's intended purpose
2. Your ideas don't have to be practical or realistic; they can be silly or strange, even, so long as they are CREATIVE uses rather than ordinary uses.
3. Avoid duplicating ideas
4. Provide brief but clear descriptions
5. The goal is to come up with creative ideas, which are ideas that strike people as clever, unusual, interesting, uncommon, humorous, innovative, or different

Please provide {num_responses} different uses, one per line, 
with no additional text or explanation.
\end{lstlisting}

\subsubsection{Test Objects}

A set of sixteen common objects was used to standardize inputs across trials. These items were selected for their general familiarity and variety in potential use contexts as used in the literature before \citep{organisciak_beyond_2023, christensen_alternate_1960}. The complete list of objects is shown in Table~\ref{tab:test-objects}.

\begin{table}[h]
\centering
\begin{tabular}{llll}
\toprule
brick & shoe & paper clip & button \\
cardboard box & pencil & bottle & newspaper \\
umbrella & pants & ball & tire \\
fork & toothbrush & & \\
\bottomrule
\end{tabular}
\caption{Test objects used in the Alternate Use Task experiments}
\label{tab:test-objects}
\end{table}

\subsubsection{Scoring and Analysis}

The experimental setup defined the model pool, the number of trials per model-object pair, and the number of responses to be generated per prompt. During each trial, the model was prompted to generate alternative uses for a specific object using one of the predefined prompt templates. Generated responses were then submitted to the OpenScoring API, which returned evaluations for originality for each response and an overall creativity score for the full response set. As a next step, a percentile ranking against human responses was calculated as a benchmark, based on $n$ = 151 from \cite{hubert_current_2024}. 

\subsection{Performed Experiments}

In the following, we provide an overview of the experiments we conducted.

\begin{enumerate}
    \item \textbf{Divergent Association Task (DAT)}: For each of the 14 LLMs, we conducted 100 independent trials using each of the two prompt variants described in Section~\ref{sec:dat-prompts}: the `DAT aware' prompt and the `DAT unaware' prompt. This resulted in a total of 2,800 DAT evaluations across all models and prompt conditions.
    
    \item \textbf{Alternate Use Task (AUT)}: For each of the 14 LLMs, we conducted 4 separate trials where each model was instructed to generate 100 alternative uses per trial. Each trial used once the `Practical and Feasible' prompt and once the `Creative and Unconventional' prompt as detailed in Section~\ref{sec:aut-prompts}. This resulted in 56 prompt-model trials, though many models, especially lower-capacity ones, generated fewer than the requested 100 uses per trial.
\end{enumerate}

All experiments were conducted under controlled conditions using standardized evaluation procedures to ensure comparability across models and prompt variations. For both tasks, we recorded the raw responses, computed scores using the respective scoring systems, and performed statistical analyses to assess performance relative to human benchmarks.

\section{Results}

Below, we first report the results for the DAT, followed by the AUT. Our main analyses focus on the DAT-awareness and the AUT-creative prompt conditions because they are more commonly used in former human studies (e.g., \citealt{reiter-palmon_scoring_2019, acar_creativity_2023}). The DAT-unawareness condition and the practical and feasible prompt condition were included for exploratory purposes. For each creativity test, we focused on percentile ranks to facilitate comparisons with human performance measures. However, we report the comparisons between humans and LLMs using the untransformed scores in the online supplemental materials on OSF\footnote{\url{https://osf.io/e62fx/?view_only=cd2f9763f84843c385ca760702eeae12}}, which replicate the results from the percentiles. Furthermore, we tested which of the LLMs performed best and how stable the responses were (i.e., computed the variability within each model).

\subsection{Divergent Association Task (DAT)}

\begin{figure}[ht]
    \centering
    \includegraphics[width=1\textwidth]{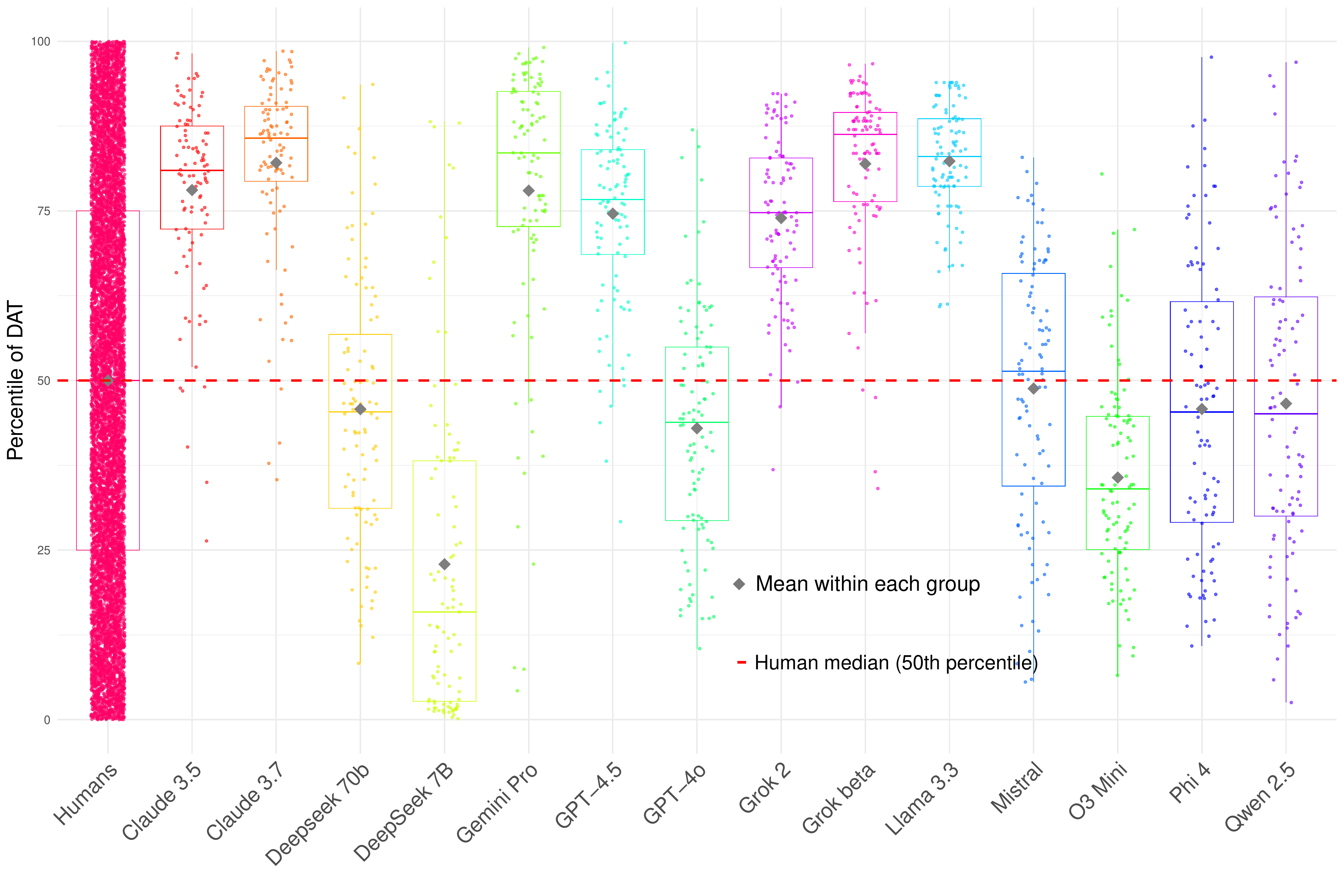}
    \caption{Percentile scores of each large language model (LLM) in the DAT-awareness condition. The first group reflects the distribution of human percentile ranks. Percentiles for the 14 LLMs were computed by benchmarking their DAT scores against the human distribution, such that higher percentiles indicate better performance relative to humans \\
    \small\textit{Note.} Each boxplot displays the distribution of percentiles for a given LLM, with black diamonds indicating mean performance. The red dashed line represents the average human performance (50th percentile). The human responses are from \cite{olson_naming_2021}}
    \label{fig:DAT_awareness_per}
\end{figure}

\begin{figure}[htbp]
    \centering
    \includegraphics[width=1\textwidth]{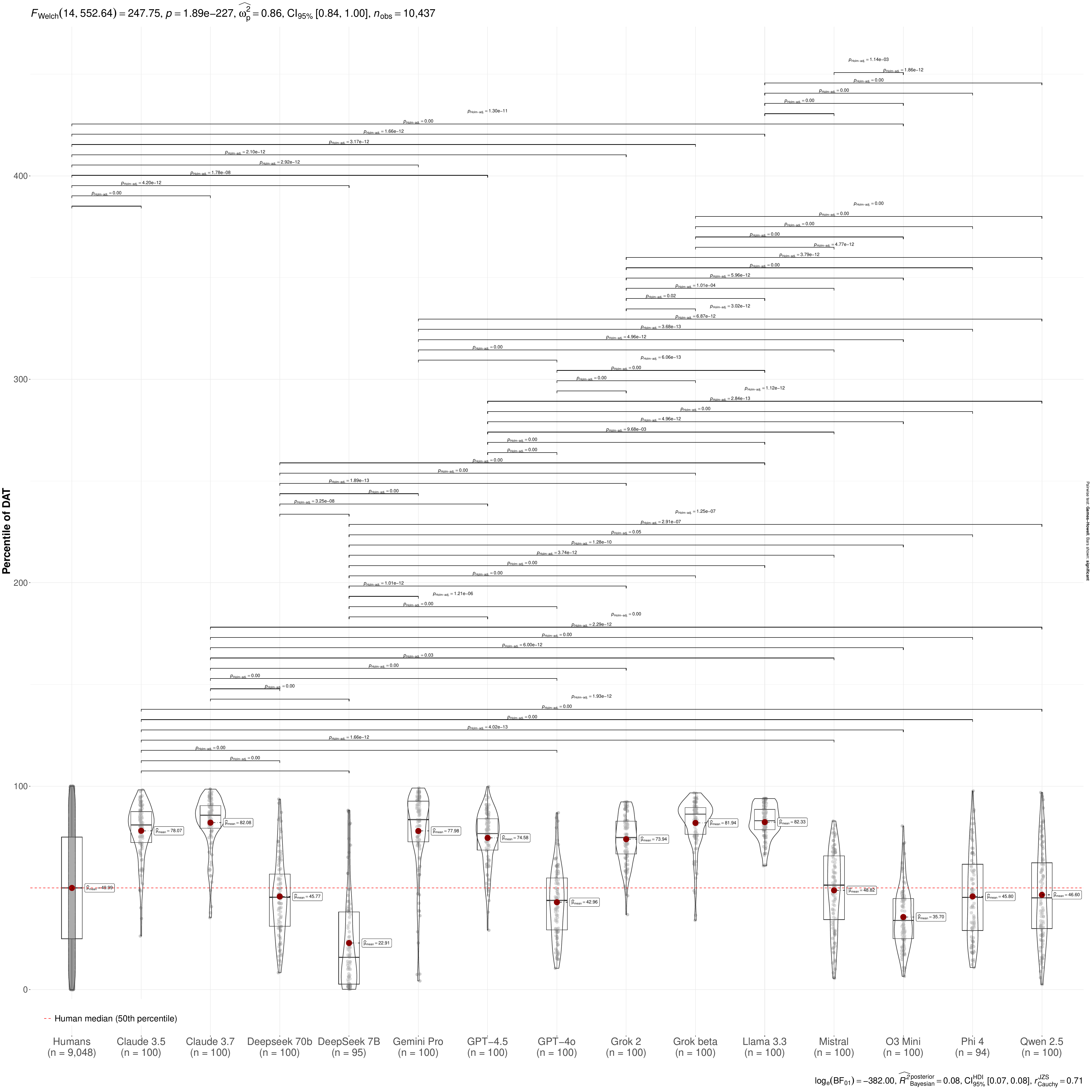}
    \caption{Percentile scores of each large language model (LLM) in the DAT-awareness. Significant pairwise comparisons between any two groups are displayed. The first group reflects the distribution of human scores.}
    \small\textit{Note.} Each boxplot displays the distribution of scores for a given LLM, with red dots indicating mean performance. The red dashed line represents the median human performance (50th percentile). The human responses are from \cite{olson_naming_2021}
    \label{fig:DAT_awareness_per_pc}
\end{figure}

The results revealed a wide spread in DAT performance between LLMs. The average percentile was with $M$~=~60.16, $SD$~=~26.15, significantly higher than the average response of humans (i.e., 50th percentile, for descriptive statistics, see Table~\ref{tab:tab_t_tests} and Figure~\ref{fig:DAT_awareness_per}), $t$~=~14.48, $p$~<~.001. Several LLMs performed, on average, poorer than the average human (i.e., a percentile rank of <50). Some models performed better than others: Llama 3.3, Claude 3.7, and Grok beta performed better than most other models and, on average, better than 80\% of humans. In contrast, DeepSeek R1 Distill performed only better than 22.91\% of humans. A pairwise comparison of models can be found in Figure \ref{fig:DAT_awareness_per_pc}.

In the next step, we compared our findings with those reported in the literature to test whether the performance of the LLMs has increased. For example, \citet{cropley_is_2023} reported in 2023 that GPT-4 had a percentile rank in the DAT of 82.54, $SD$~=~13.94 across 102 responses. Surprisingly, an independent sample t-test revealed that GPT-4o performed worse at the end of February 2025 (i.e., in our data) than GPT-4 in the data from Cropley in 2023, $t$(200)~=~14.46, $p$~<~.001, $d$~=~2.04. Further, \citet{hubert_current_2024} reported that GPT-4 had a score (no percentile ranks were reported) of $M$~=~84.56, $SD$~=~3.05, across 151 responses, also in 2023. This was again higher than the scores we found in February 2025, $M$~=~77.34, $SD$ = 2.92, $t$(249)~=~18.84, $p$~<~.001, $d$~=~2.41.

\begin{figure}[htbp]
    \centering
    \includegraphics[width=1\textwidth]{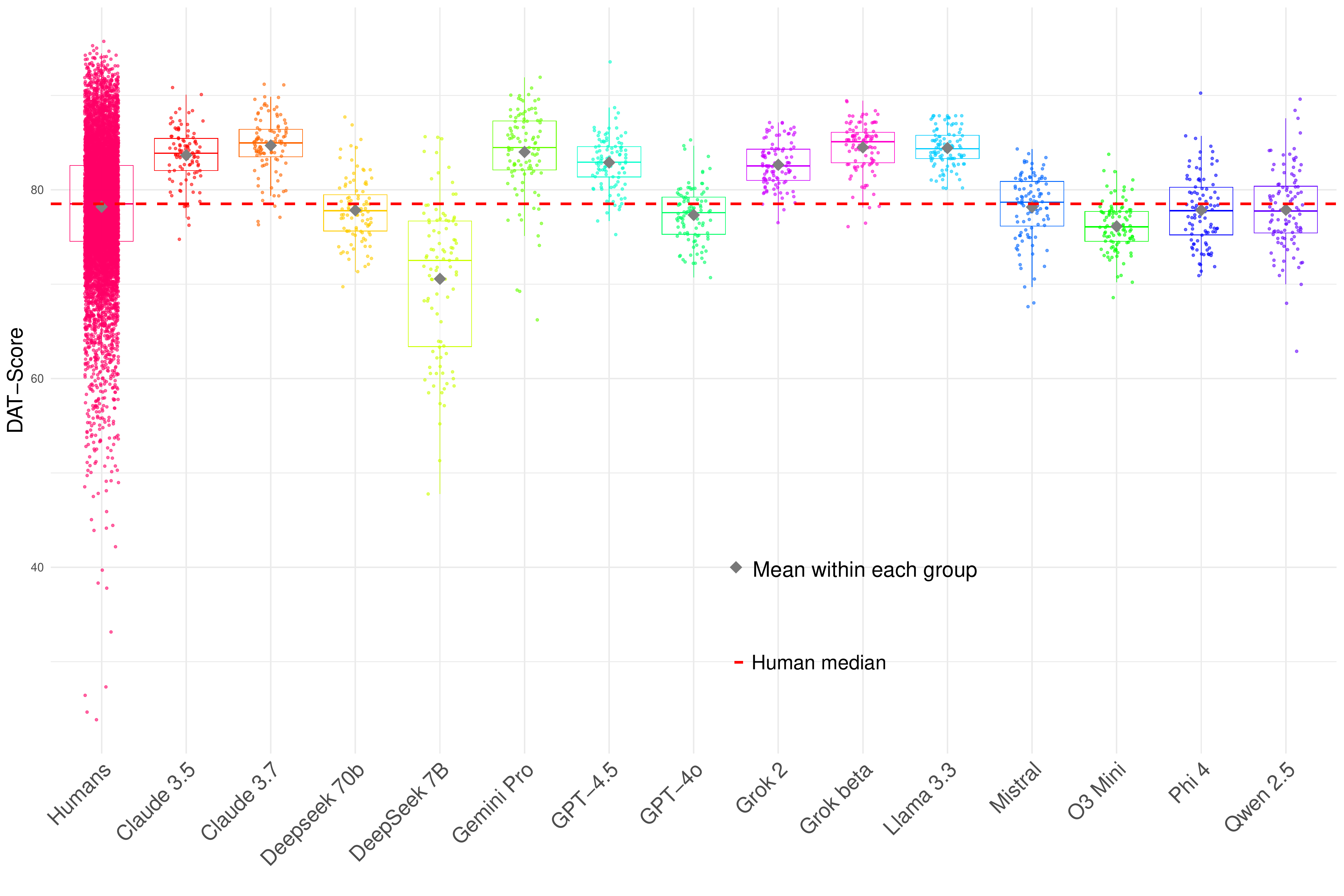}
    \caption{Scores of each large language model (LLM) in the DAT-awareness condition. The first group reflects the distribution of human scores.}
    \small\textit{Note.} Each boxplot displays the distribution of scores for a given LLM, with black diamonds indicating mean performance. The red dashed line represents the median human performance. The human responses are from \cite{olson_naming_2021}
    \label{fig:DAT_awareness_scores}
\end{figure}

\begin{figure}[htbp]
    \centering
    \includegraphics[width=1\textwidth]{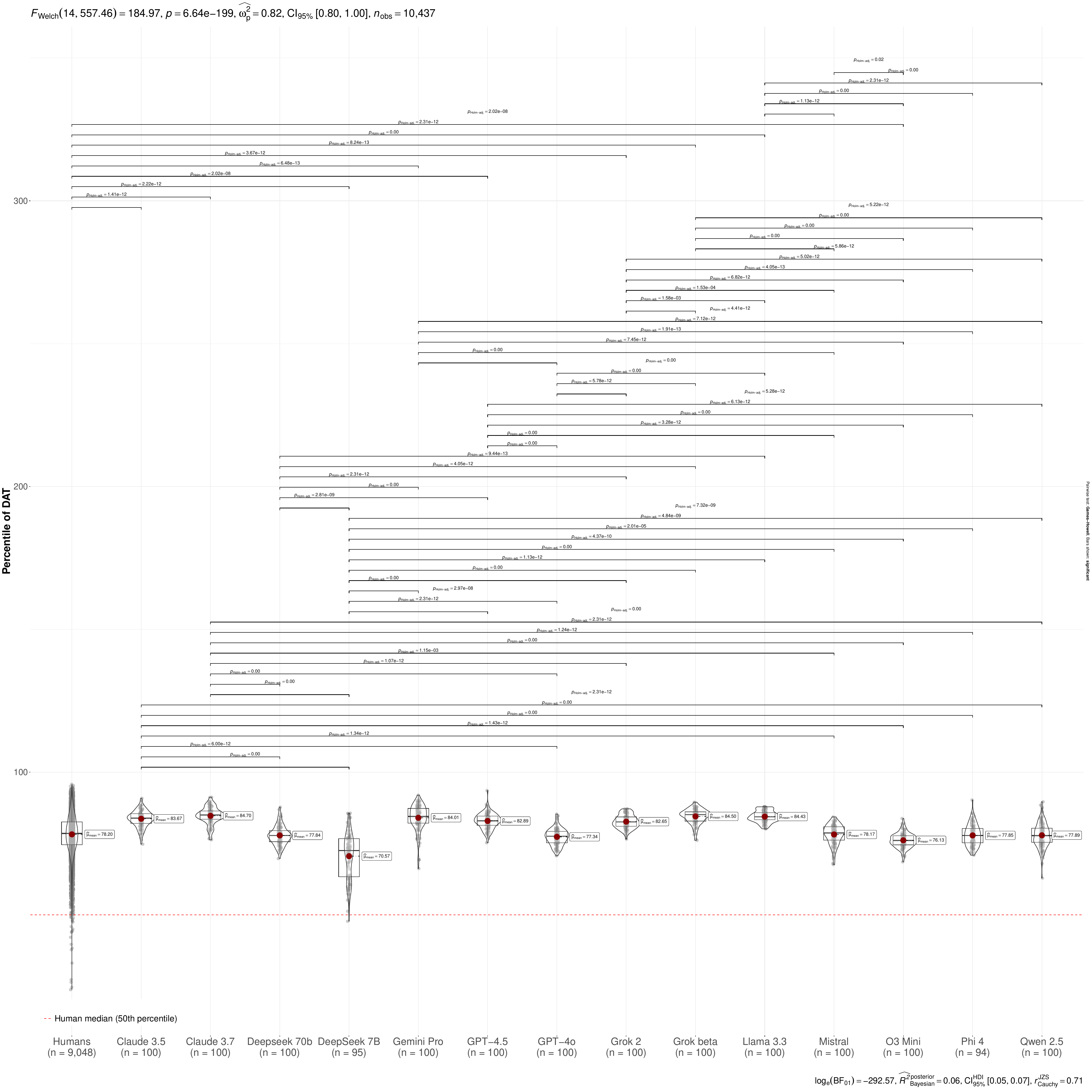}
    \caption{Scores of each large language model (LLM) in the DAT-awareness condition. Significant pairwise comparisons between any two groups are displayed. The first group reflects the distribution of human scores.}
    \small\textit{Note.} Each boxplot displays the distribution of scores for a given LLM, with red dots indicating mean performance. The red dashed line represents the median human performance. The human responses are from \cite{olson_naming_2021}
    \label{fig:DAT_awareness_scores_pc}
\end{figure}

However, the variability within the responses was substantial for almost all LLMs (cf. Figure~\ref{fig:DAT_awareness_per}). Across all 14 LLMs, 495 tests were below the 50th percentile and 894 were above the 50th percentile, \textit{range}~=~0.16, 99.78. The only LLM that produced consistent responses above the average human (i.e., >50th percentile) was Llama. Even for LLMs that are considered powerful such as ChatGPT-4.5, 6 responses were below the 50th percentile, although 94 were above the 50th percentile ($M$~=~74.58, $SD$~=~13.45, Table \ref{tab:tab_t_tests}). Since transforming scores into percentiles can affect the distance between scores, we also compared raw scores, which replicated the findings reported in this section (see Figures \ref{fig:DAT_awareness_scores} and \ref{fig:DAT_awareness_scores_pc}).

\begin{figure}[htbp]
    \centering
    \includegraphics[width=1\textwidth]{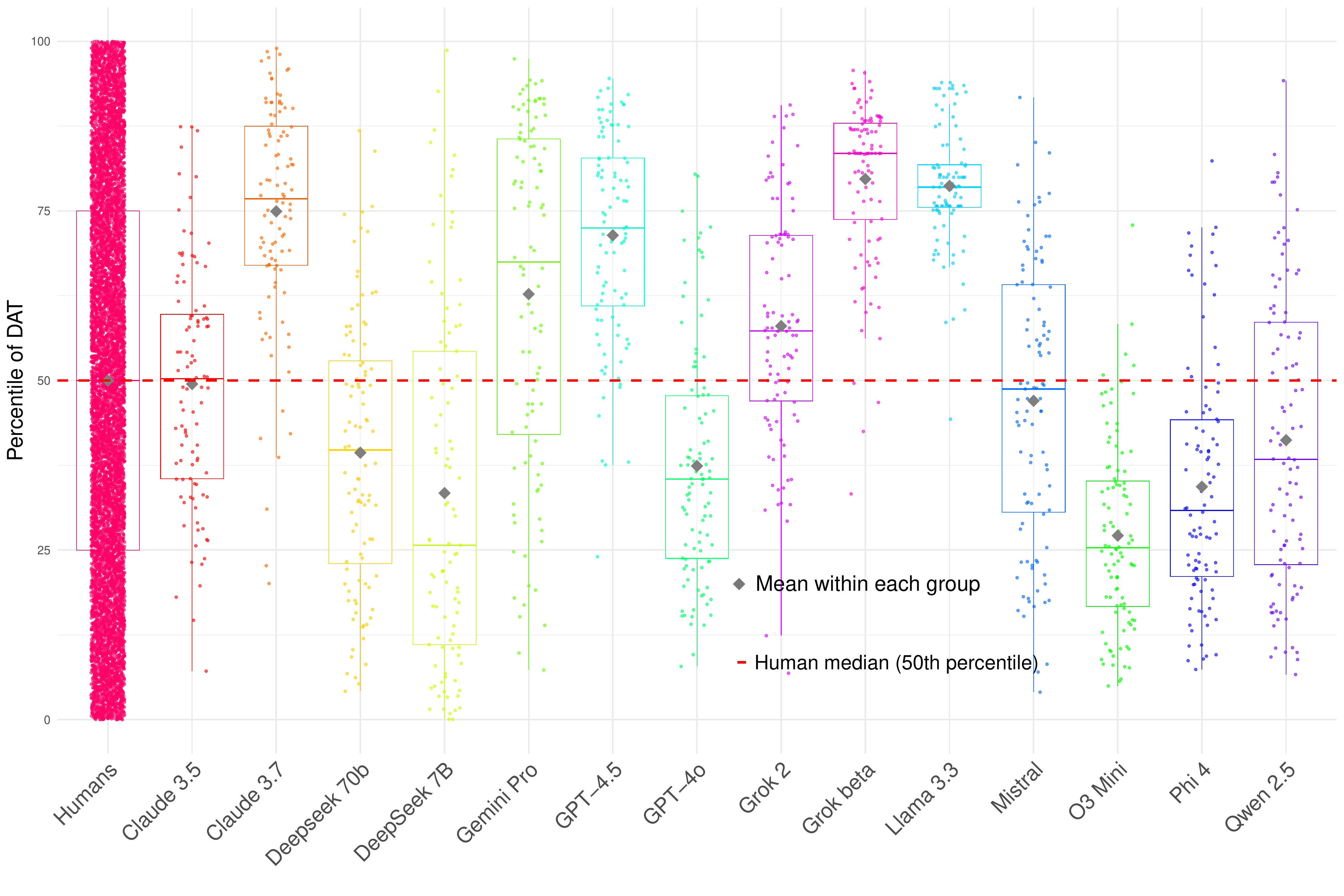}
    \caption{Percentile scores of each large language model (LLM) in the DAT-unawareness condition. The first group reflects the distribution of human percentile ranks. Percentiles for the 14 LLMs were computed by benchmarking their DAT scores against the human distribution, such that higher percentiles indicate better performance relative to humans \\
    \small\textit{Note.} Each boxplot displays the distribution of percentiles for a given LLM, with black diamonds indicating mean performance. The red dashed line represents the average human performance (50th percentile). The human responses are from \cite{olson_naming_2021}}
    \label{fig:DAT_unawareness_per}
\end{figure}

\begin{figure}[htbp]
    \centering
    \includegraphics[width=1\textwidth]{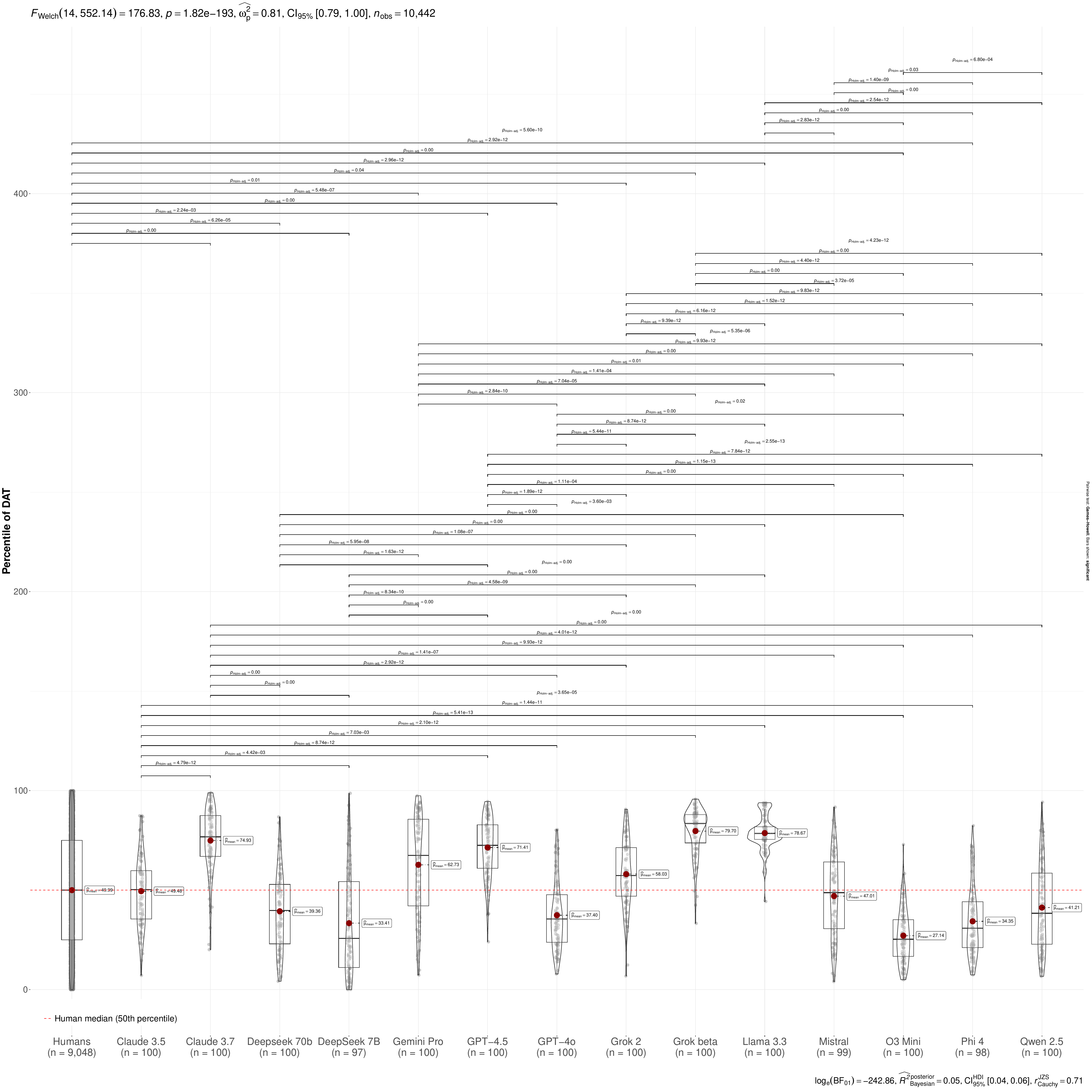}
    \caption{Percentile scores of each large language model (LLM) in the DAT-unawareness condition. Significant pairwise comparisons between any two groups are displayed. The first group reflects the distribution of human scores.}
    \small\textit{Note.} Each boxplot displays the distribution of scores for a given LLM, with red dots indicating mean performance. The red dashed line represents the median human performance (50th percentile). The human responses are from \cite{olson_naming_2021}
    \label{fig:DAT_unawareness_per_pc}
\end{figure}

\begin{figure}[htbp]
    \centering
    \includegraphics[width=1\textwidth]{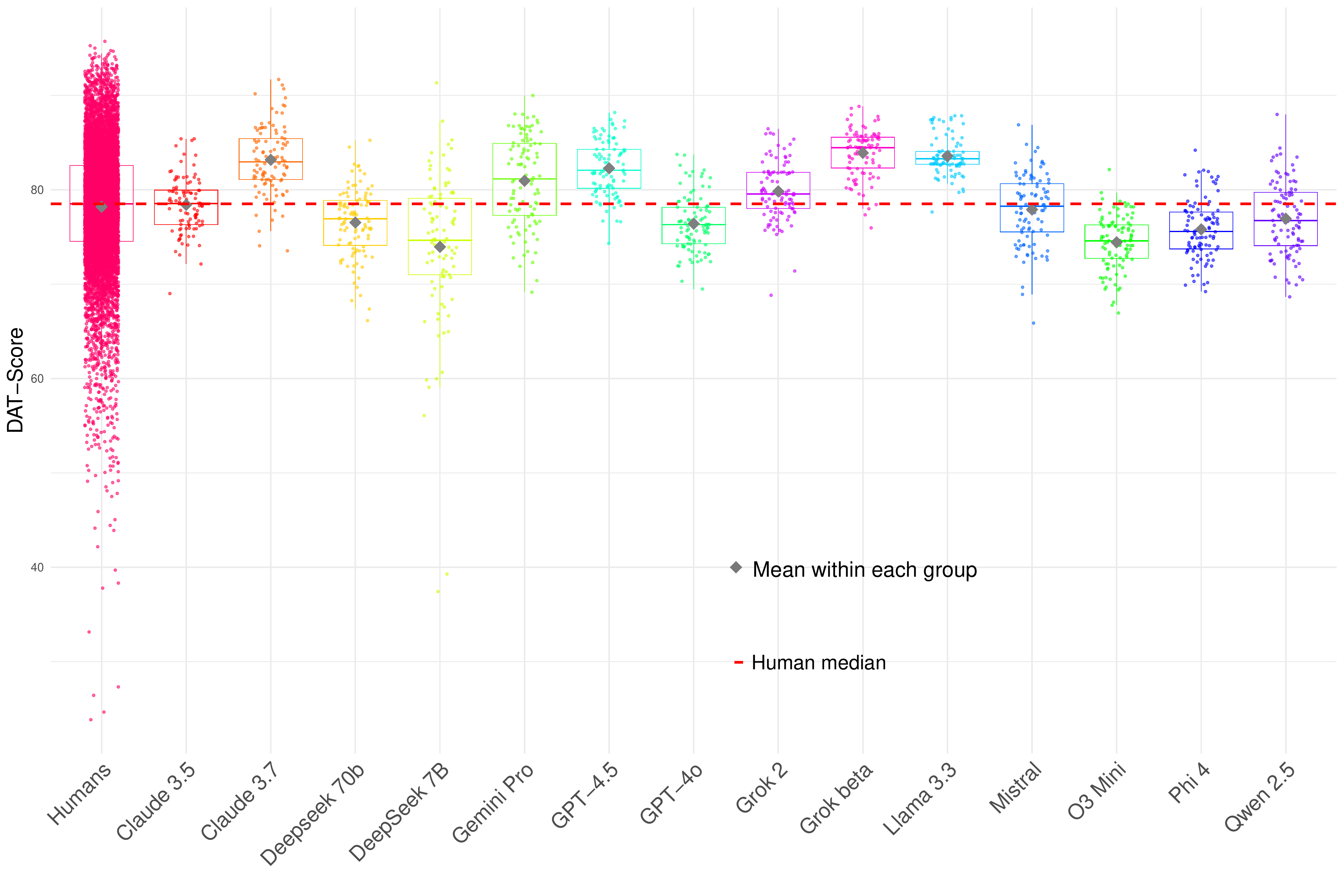}
    \caption{Scores of each large language model (LLM) in the DAT-unawareness condition. The first group reflects the distribution of human scores.}
    \small\textit{Note.} Each boxplot displays the distribution of scores for a given LLM, with black diamonds indicating mean performance. The red dashed line represents the median human performance. The human responses are from \cite{olson_naming_2021}
    \label{fig:DAT_unawareness_scores}
\end{figure}

\begin{figure}[htbp]
    \centering
    \includegraphics[width=1\textwidth]{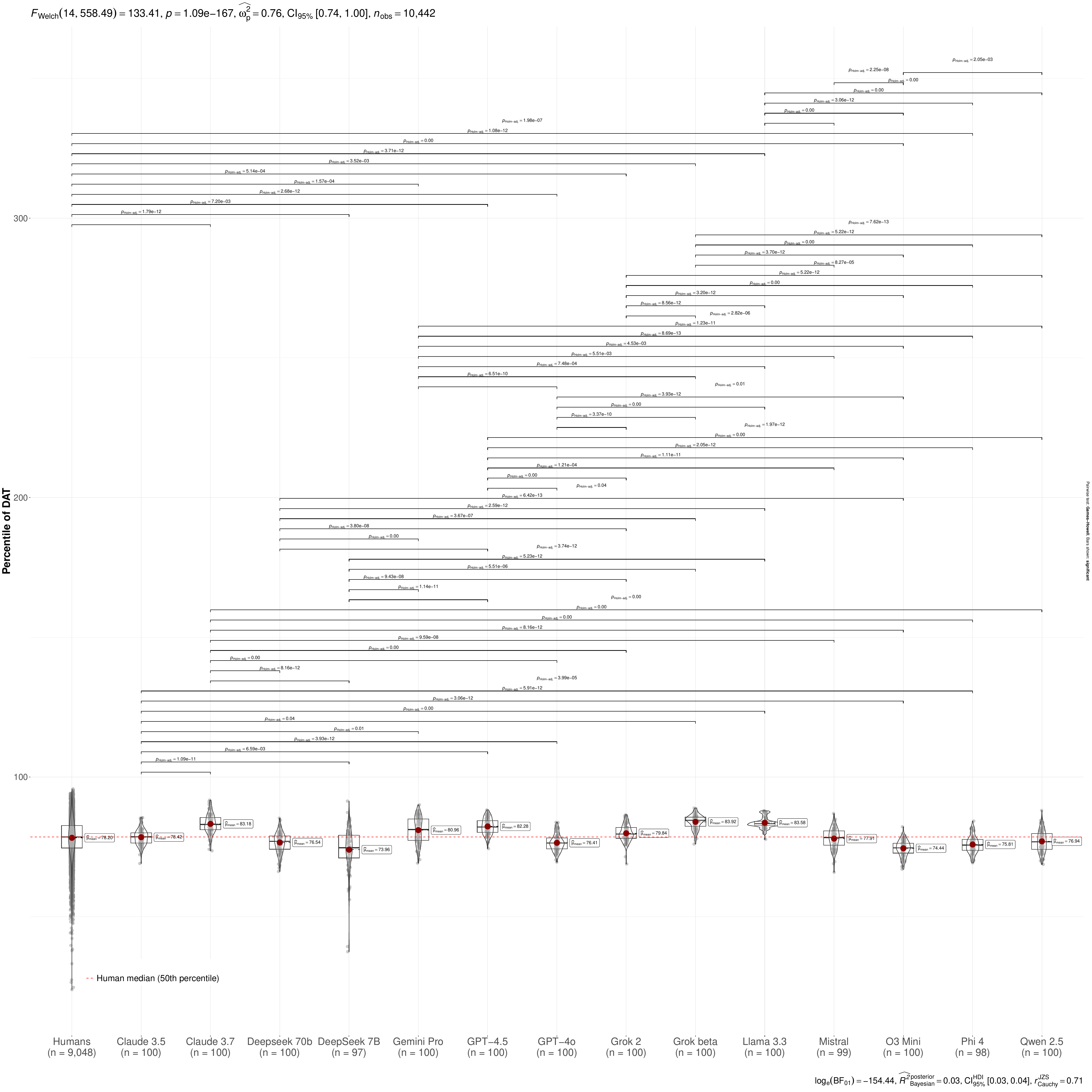}
    \caption{Scores of each large language model (LLM) in the DAT-unawareness condition. Significant pairwise comparisons between any two groups are displayed. The first group reflects the distribution of human scores.}
    \small\textit{Note.} Each boxplot displays the distribution of scores for a given LLM, with red dots indicating mean performance. The red dashed line represents the median human performance. The human responses are from \cite{olson_naming_2021}
    \label{fig:DAT_unawareness_scores_pc}
\end{figure}

Additionally, we tested whether mentioning the DAT in the prompt matters. We compared the 100 responses per LLM in the DAT-aware condition with the 100 responses per LLM in the DAT-unaware condition using a linear mixed-effects model (R packages lme4 and lmerTest; \citealt{kuznetsova_lmertest_2017}), specifying random intercepts for the 14 LLMs. On average, the LLMs in the aware condition scored higher ($M$ = 60.16, $SD$ = 26.15) than in the unaware condition ($M$ = 52.56, $SD$ = 25.38), $B$ = 11.89, $SE$ = 0.90, $p$ < .001. Interestingly, exploratory follow-up analyses revealed that this effect differed between LLMs: Claude 3.5 and Grok 2 performed much better in the DAT-aware condition (Cohen's ds > 1.00), whereas DeepSeekR1 Distill Qwen 7B performed significantly worse, with most other models performing somewhat better in the aware condition (cf. Table \ref{tab:tab_t_tests}). Note that we only interpret findings that are significant at $\alpha$ = .005 to adjust for the 14 comparisons (for visualizations and pairwise comparisons between the models, see Figures \ref{fig:DAT_unawareness_per}, \ref{fig:DAT_unawareness_per_pc}, \ref{fig:DAT_unawareness_scores}, and \ref{fig:DAT_unawareness_scores_pc}).

\begin{table}[htbp]
\caption{Comparison of conditions in which prompt mentioned DAT vs. not mentioned}
\centering
\resizebox{\textwidth}{!}{%
\begin{tabular}[t]{lrrrrrrrrrrr}
\toprule
& \multicolumn{4}{c}{With DAT-awareness} & \multicolumn{4}{c}{Without DAT-awareness} & & & \\
Model & M & SD & < 50 & > 50 & M & SD & < 50 & > 50 & t-value & p-value & Cohen's d \\
\midrule
Claude 3.5 & 78.07 & 13.78 & 6 & 94 & 49.48 & 17.44 & 49 & 51 & 12.86 & <.001 & 1.82\\
Claude 3.7 & 82.08 & 13.35 & 4 & 96 & 74.93 & 16.48 & 7 & 93 & 3.37 & 0.001 & 0.48\\
DeepSeek R1 70b & 45.77 & 19.45 & 66 & 34 & 39.36 & 19.65 & 69 & 31 & 2.32 & 0.021 & 0.33\\
DeepSeek R1 Distill Qwen 7B & 22.91 & 23.70 & 84 & 11 & 33.41 & 26.15 & 70 & 27 & -2.91 & 0.004 & -0.42\\
Gemini Pro & 77.98 & 20.76 & 11 & 89 & 62.73 & 25.51 & 34 & 66 & 4.64 & <.001 & 0.66\\
o3 Mini & 35.70 & 14.81 & 86 & 14 & 27.14 & 13.67 & 95 & 5 & 4.24 & <.001 & 0.60\\
GPT-4o & 42.96 & 17.35 & 67 & 33 & 37.40 & 17.27 & 79 & 21 & 2.27 & 0.024 & 0.32\\
GPT-4.5 & 74.58 & 13.56 & 6 & 94 & 71.41 & 15.13 & 8 & 92 & 1.56 & 0.120 & 0.22\\
Llama 3.3 70B & 82.33 & 8.25 & 0 & 100 & 78.67 & 8.64 & 1 & 99 & 3.06 & 0.002 & 0.43\\
Mistral Nemo Instruct 2407 & 48.82 & 19.63 & 45 & 55 & 47.01 & 20.76 & 56 & 43 & 0.63 & 0.527 & 0.09\\
Phi 4 & 45.80 & 21.60 & 57 & 37 & 34.35 & 17.73 & 81 & 17 & 4.02 & <.001 & 0.58\\
Qwen 2.5 7B Instruct 1M & 46.60 & 22.97 & 56 & 44 & 41.21 & 21.71 & 63 & 37 & 1.70 & 0.090 & 0.24\\
Grok 2 & 73.94 & 11.73 & 3 & 97 & 58.03 & 17.30 & 33 & 67 & 7.62 & <.001 & 1.08\\
Grok beta & 81.94 & 12.22 & 4 & 96 & 79.70 & 12.02 & 4 & 96 & 1.31 & 0.193 & 0.18\\
\bottomrule
\end{tabular}%
}
\label{tab:tab_t_tests}
\end{table}

\subsection{Alternate Use Task (AUT)}
\begin{figure}[ht]
    \centering
    \includegraphics[width=1\textwidth]{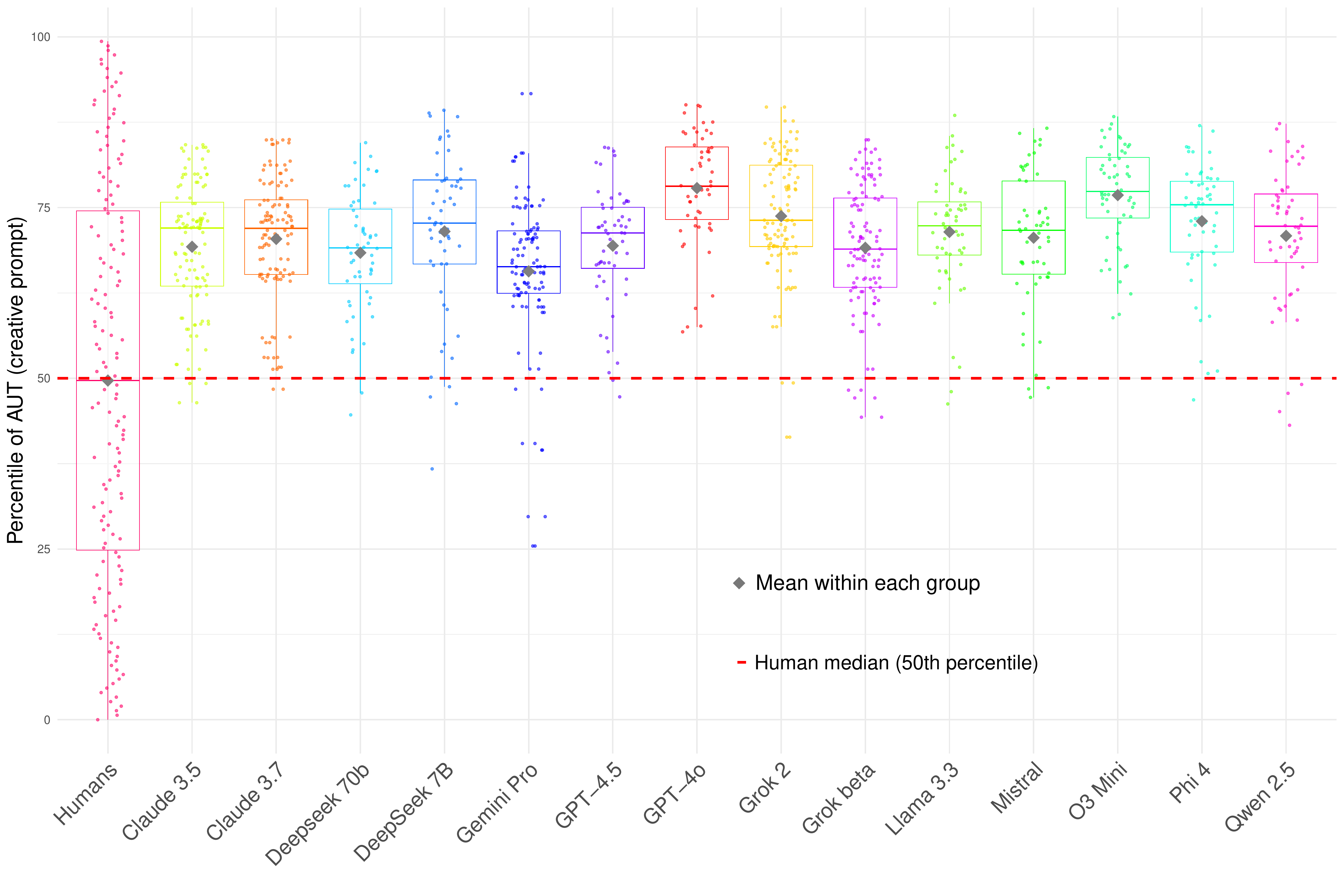}
    \caption{Percentile scores of each large language model (LLM) in the creative-prompt condition (AUT). The first group reflects the distribution of human percentile ranks. Percentiles for the 14 LLMs were computed by benchmarking their AUT scores against the human distribution, such that higher percentiles indicate better performance relative to humans \\
    \small\textit{Note.} Each boxplot displays the distribution of percentiles for a given LLM, with black diamonds indicating mean performance. The red dashed line represents the average human performance (50th percentile). The human responses are from \cite{hubert_current_2024}}    \label{fig:AUT_creative}
\end{figure}

\begin{figure}[htbp]
    \centering
    \includegraphics[width=1\textwidth]{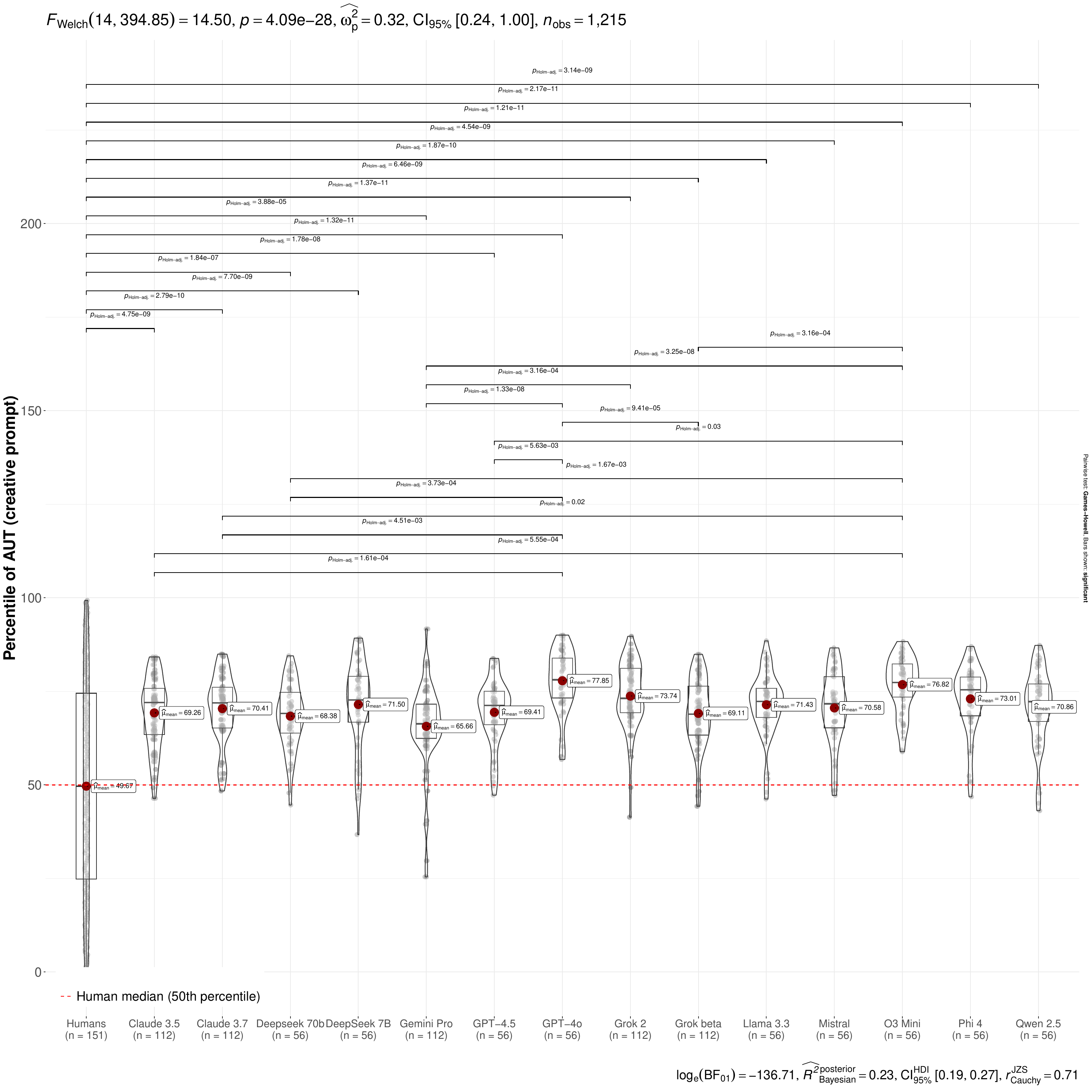}
    \caption{Percentile scores of each large language model (LLM) in the practical-prompt condition (AUT). Significant pairwise comparisons between any two groups are displayed. Percentiles for the 14 LLMs were computed by benchmarking their AUT scores against the human distribution, such that higher percentiles indicate better performance relative to humans. The first group reflects the distribution of human scores.}
    \small\textit{Note.} Each boxplot displays the distribution of scores for a given LLM, with red dots indicating mean performance. The red dashed line represents the median human performance (50th percentile). The human responses are from \cite{hubert_current_2024}
    \label{fig:AUT_creative_pc}
\end{figure}

Again, we focused on the percentiles to facilitate comparisons with human responses. A one-way between-subjects ANOVA with 14 levels was significant, $F(13, 356.64)~=~9.34, p~<~.001$. Holm-adjusted follow-up tests revealed that several of the models performed differently from each other, with GPT-4o performing overall best and Gemini Pro relatively worst (Figure \ref{fig:AUT_creative}; for pairwise comparisons between the 14 LLMs and human responses, see Figure \ref{fig:AUT_creative_pc}). Overall, the average model performance was more homogeneous than for the DAT, with all means ranging between 65.66 and 77.85 (Table \ref{tab:AUT}). A series of one-sample t-tests revealed that each model performed significantly better than the average human (i.e., percentile of 50), $p$s~<~.001.

In the next step, we compared our findings with those reported in the literature to test whether the performance of the LLMs has increased. For example, \citet{haase_artificial_2023} found that the average originality scores generated by GPT-4 in March 2023 for the prompts \textit{pants, ball, tire, fork,} and \textit{toothbrush} was $M~=~3.22, SD~=~0.36$, as analyzed with OpenScoring \citep{organisciak_beyond_2023}. This was not significantly different from the average in our sample, $M~=~3.67, SD~=~0.27, t(7.44)~=~2.23, p=~.059, d~=~1.41$. 

We again found heterogeneity in the responses. The best response in each model was 25 percentile points (i.e., one quartile) higher than the model's worst response (Figure~\ref{fig:AUT_creative}). Somewhat surprisingly, most responses were below the top 10\% of human-generated responses. Only 3 out of 1,061 responses, 0.28\%, were in the top 10\%. 

\begin{figure}[htbp]
    \centering
    \includegraphics[width=1\textwidth]{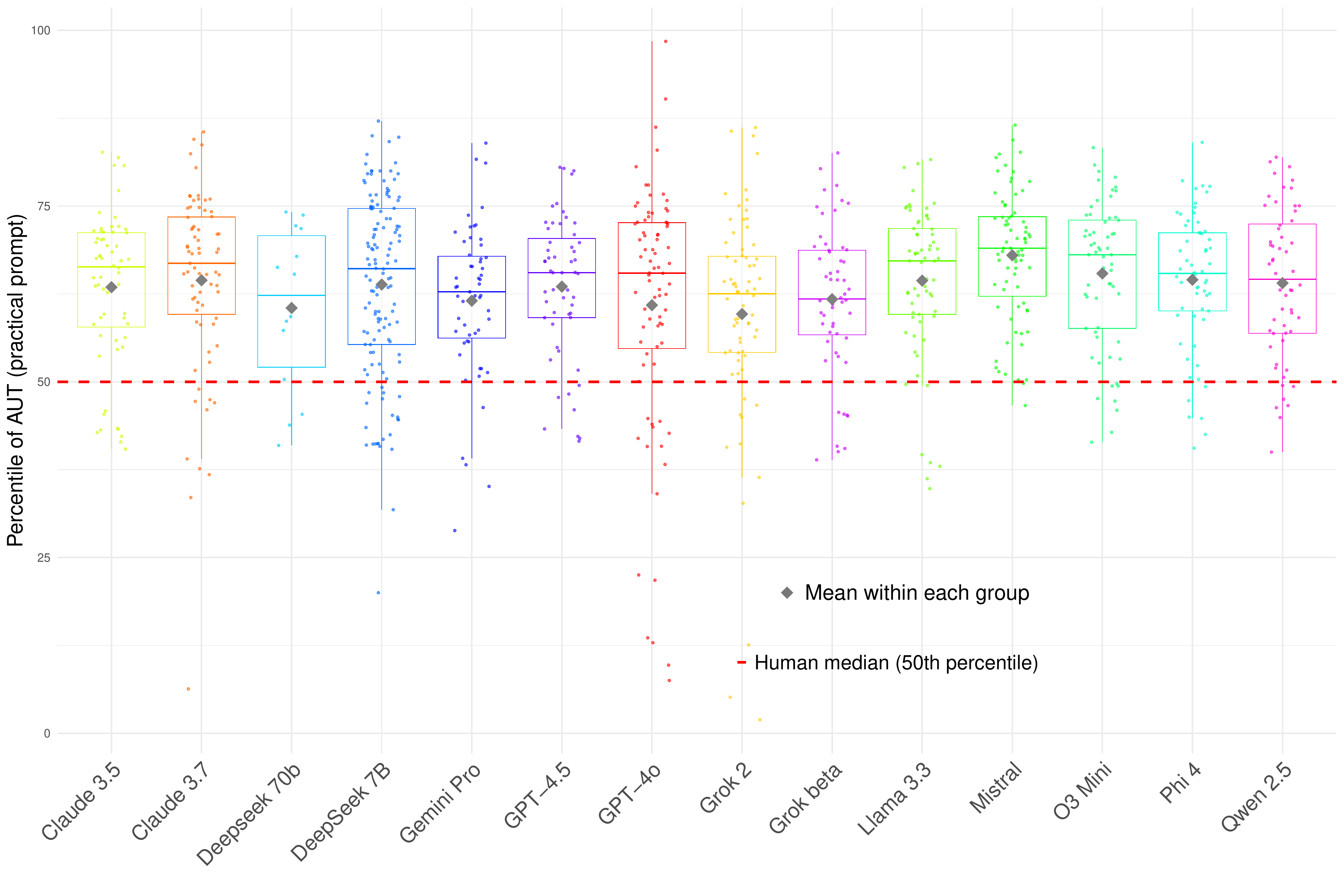}
    \caption{Percentile scores of each large language model (LLM) in the practical-prompt condition (AUT). Percentiles for the 14 LLMs were computed by benchmarking their AUT scores against the human distribution, such that higher percentiles indicate better performance relative to humans \\
    \small\textit{Note.} Each boxplot displays the distribution of percentiles for a given LLM, with black diamonds indicating mean performance. The red dashed line represents the average human performance (50th percentile). The human responses are from \cite{hubert_current_2024}}    
    \label{fig:AUT_practical}
\end{figure}

\begin{figure}[htbp]
    \centering
    \includegraphics[width=1\textwidth]{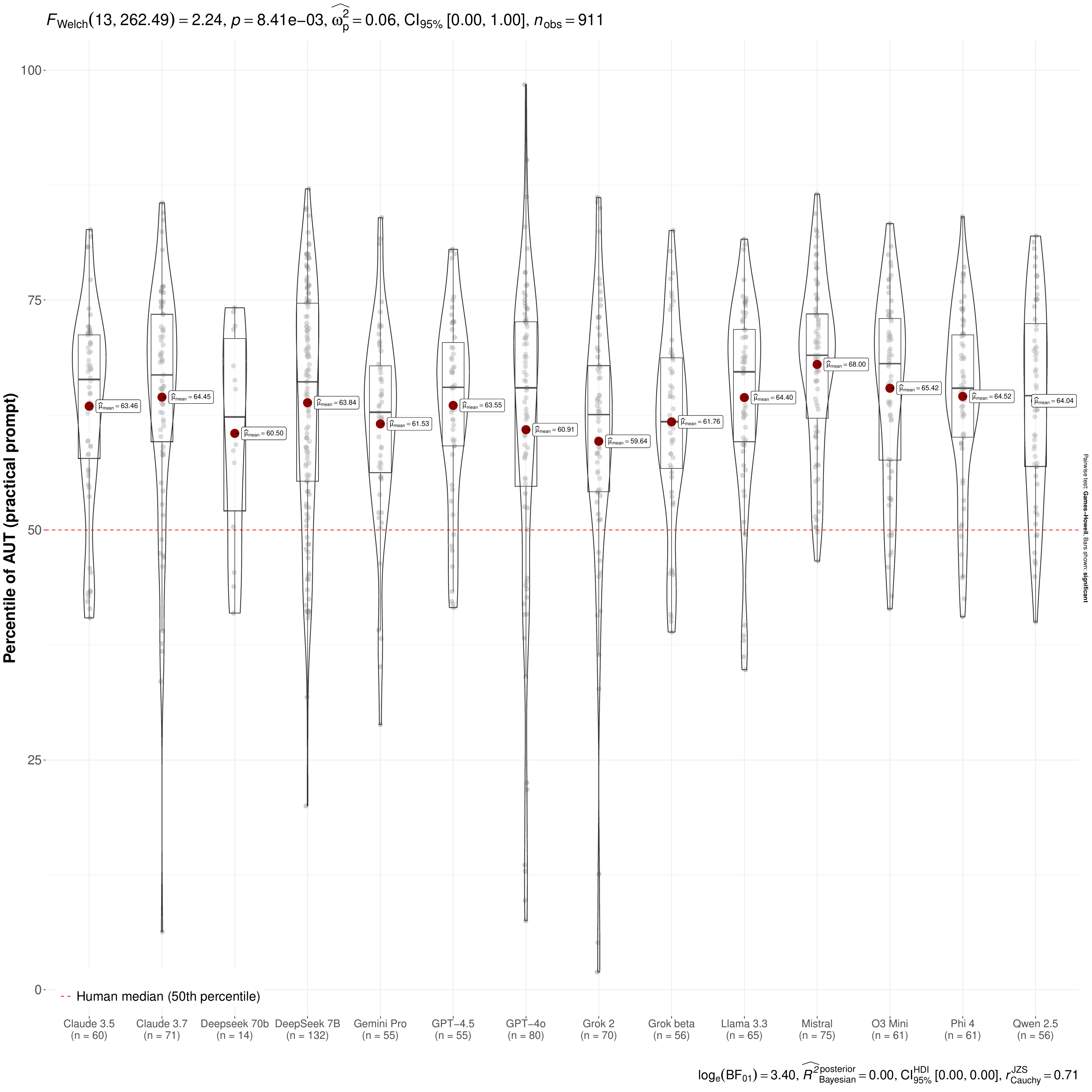}
    \caption{Percentile scores of each large language model (LLM) in the practical-prompt condition (AUT). Percentiles for the 14 LLMs were computed by benchmarking their AUT scores against the human distribution, such that higher percentiles indicate better performance relative to humans. }
    \small\textit{Note.} Each boxplot displays the distribution of scores for a given LLM, with red dots indicating mean performance. The red dashed line represents the median human performance (50th percentile). The human responses are from \cite{hubert_current_2024}. No two LLMs differ significantly from each other after Holm-corrections.
    \label{fig:AUT_practical_pc}
\end{figure}

Additionally, we tested whether prompting the LLMs to be more creative (vs practical and feasible) would result in more original responses using again a linear mixed-effects model with random intercepts across the 14 LLMs. On average, LLMs in the creative condition scored higher ($M$ = 70.85, $SD$ = 10.00) than LLMs in the practical condition ($M$ = 63.49, $SD$ = 12.45), $B$ = 7.89, $SE$ = 0.51, $p$ < .001. Exploratory follow-up analyses revealed that this effect was mostly consistent across LLMs: All LLMs performed better when instructed to be creative (vs practical), even though this effect was not significant for Mistral (Figures \ref{fig:AUT_practical} and \ref{fig:AUT_practical_pc} for visualisations of the scores from the practical prompt condition).

\begin{table}[ht]
\centering
\caption{Comparisons between practical and creative prompts, separately for each of the 14 large language models}
\begin{tabular}[t]{lrrrrrrrrrrr}
\toprule
& \multicolumn{4}{c}{Practical prompt} & \multicolumn{4}{c}{Creative prompt} & & & \\
Model & M & SD & < 50 & > 50 & M & SD & < 50 & > 50 & t-value & p-value & Cohen's d\\
\midrule
GPT-4o & 60.91 & 18.05 & 16 & 64 & 77.85 & 8.32 & 0 & 56 & -6.55 & <.001 & -1.14\\
Claude 3.7 & 64.45 & 13.42 & 10 & 61 & 70.41 & 9.13 & 2 & 110 & -3.58 & <.001 & -0.54\\
Grok 2 & 59.64 & 15.82 & 11 & 59 & 73.74 & 9.21 & 4 & 108 & -7.60 & <.001 & -1.16\\
Claude 3.5 & 63.46 & 10.88 & 9 & 51 & 69.26 & 9.45 & 4 & 108 & -3.63 & <.001 & -0.58\\
Llama 3.3 & 64.40 & 10.81 & 7 & 58 & 71.43 & 8.61 & 2 & 54 & -3.91 & <.001 & -0.71\\
\addlinespace
Mistral & 68.00 & 9.25 & 2 & 73 & 70.58 & 9.86 & 3 & 53 & -1.54 & 0.127 & -0.27\\
o3 Mini & 65.42 & 10.38 & 8 & 53 & 76.82 & 7.30 & 0 & 56 & -6.82 & <.001 & -1.26\\
Phi 4 & 64.52 & 9.70 & 6 & 55 & 73.01 & 8.98 & 1 & 55 & -4.90 & <.001 & -0.91\\
DeepSeek 70b & 60.50 & 11.58 & 3 & 11 & 68.38 & 8.71 & 2 & 54 & -2.83 & 0.006 & -0.84\\
DeepSeek 7B & 63.84 & 13.01 & 22 & 110 & 71.50 & 12.01 & 4 & 52 & -3.77 & <.001 & -0.60\\
\addlinespace
Gemini Pro & 61.53 & 10.94 & 5 & 50 & 65.66 & 11.77 & 10 & 102 & -2.18 & 0.031 & -0.36\\
GPT-4.5 & 63.55 & 10.22 & 8 & 47 & 69.41 & 8.65 & 2 & 54 & -3.27 & 0.002 & -0.62\\
Grok beta & 61.76 & 10.53 & 8 & 48 & 69.11 & 9.39 & 6 & 106 & -4.59 & <.001 & -0.75\\
Qwen 2.5 & 64.04 & 10.69 & 7 & 49 & 70.86 & 10.07 & 4 & 52 & -3.47 & 0.001 & -0.66\\
\bottomrule
\end{tabular}
\label{tab:AUT}
\end{table}

\section{Discussion}

In the present study, we tested whether LLMs have become more creative over time across two commonly used creativity tests, assessed the variability of their responses, and examined performance differences across 14 widely used models. Regarding the first research question, we found that GPT-4o --previously benchmarked in 2023 as GPT-4-- performed substantially worse on the Divergent Association Task (DAT) but retained its performance on the Alternative Uses Task (AUT). Even for the AUT, however, only 0.28\% of responses reached the 90\textsuperscript{th} percentile. In other words, highly creative responses remain rare, and humans are still approximately 35.7 times more likely to produce such standout ideas. This finding offers one possible explanation for the increasingly documented trend toward homogenization in LLM-assisted output \citep{doshi_generative_2024, moon_homogenizing_2024, anderson_homogenization_2024}. While LLMs may generate text that appears individually novel, they often lack the type of originality required to break into the top decile of human creativity.

Interestingly, we found substantial differences between models in the DAT, whereas performance on the AUT was, on average, higher and more consistent across models. One explanation may be that the AUT, being widely used in GenAI research, is overrepresented in training data or has influenced model optimization, leading to inflated and stable performance. The DAT, by contrast, is less common and structurally more demanding: it requires generating a fixed number of semantically distant words, a task that combines lexical control with abstract association. This may be harder for some models to interpret, especially given sparse prompt cues and their tendency to favor locally coherent outputs. Further, such constraints may be less compatible with the default autoregressive generation mode of LLMs, particularly if the prompt does not provide sufficient context or examples. Recent evidence suggests that LLMs can struggle with tasks that require generation under sparse or underspecified instructions \citep{petrov2025proof}.

We further observed substantial within-model variability, especially in DAT performance. Even the same model under the same conditions often produced responses ranging from below-average to exceptional. This intra-model instability complicates the evaluation of LLM creativity. While prior studies often relied on single-shot assessments or few-shot comparisons, our findings suggest that such designs may over- or underestimate the actual creative potential of these systems. The variability we observed exceeds that previously reported in LLM evaluations focused on structured or closed-ended tasks (e.g., logical reasoning, legal analysis, see \citealp{atil_llm_2024, blair2025llms, liu2024your}), likely because divergent thinking tasks invite broader exploration and lower constraints by design.

This variability carries important implications for both creativity research and practical applications. It may help explain the inconsistent findings in studies exploring human-AI co-creativity, where collaboration with LLMs sometimes enhances and sometimes inhibits creative performance (cf. Table~\ref{tab:creativity_overview}; \citealt{vaccaro_when_2024, taheri_investigating_2024}). As our results show, creativity outcomes are highly sensitive to model choice, prompt framing, and stochastic model behavior. Without accounting for these factors, conclusions about LLM effectiveness in creative domains may be premature or misleading.

Prompt design emerged as another significant modulator of performance. We found that merely disclosing the creative test context (e.g., mentioning the DAT) influenced LLM performance in model-specific ways --improving results for some models (e.g., Claude 3.5 and Grok 2), while worsening performance for DeepSeek R1 Distill Qwen 7B. This aligns with recent findings showing that LLMs are sensitive to goal framing and task specification \citep{memmert_more_2024}, and echoing human creativity research on priming effects, where some individuals perform better when prompted to ``be creative'' \citep{sassenberg_dont_2005, acar_what_2020}. The implication is that creativity in LLMs may, in part, be prompt-contingent -- a result of interaction dynamics rather than an inherent capacity.

Our findings also speak to the larger philosophical debate on \textit{artificial creativity}. Critics argue that LLMs merely remix existing data, lacking the emotional depth, intentionality, or conceptual leaps characteristic of human creativity \citep{runco_ai_2023, cropley_creativity_2023}. Indeed, the absence of high-end originality in LLM output could be taken as support for this view. However, we caution against such binary thinking. While LLMs may not engage in creative processes in the human sense, their ability to generate outputs that score above the average human in both semantic divergence and usefulness indicates a form of functional or output-based creativity. Beyond all critical analysis discussed so far, 80\% of LLMs AUT-output is on average better than that of humans -- when specifically picking a ``creative LLM'', like Claude, Gemini, GPT-4o, or Grok, one can expect original output. 

What these results lead to, especially the variability, is the importance of actively working with and reflecting upon the output generated by LLMs. Thus, our results contribute to the growing literature on LLMs in human-AI co-creativity. While LLMs clearly offer utility as brainstorming aids or creative scaffolds, their performance is not reliably high nor uniformly distributed across contexts. The most promising use cases may lie not in full automation but in human-AI teaming, where the human provides context and framing, and the LLM contributes fluent, varied, or unexpected content. Especially the LLMs' capabilities to be extremely fast, easy to acces,s and very fluent in output length makes them very efficient co-creators. Still, as our findings show, such systems may encourage mid-level novelty but rarely produce radically original ideas --thus reinforcing combinatorial rather than conceptual creativity \citep{soroush_creativity_2025, orru_human-like_2023}. Without thoughtful human oversight and critical engagement, GenAI may unintentionally constrain creative diversity and reinforce existing patterns, rather than expanding the overall human-AI-enhanced creative process.

In sum, while LLMs demonstrate measurable creative potential and even outperform average human performance in certain contexts, substantial variability across and within models calls for conscious interaction with models and their output. The evaluation of LLM creativity must consider both methodological design and practical framing --especially if these tools are to be integrated meaningfully into human creative workflows.

\subsection{Limitations and Ethical Considerations}

This study has several limitations. First, we used the default settings provided by each LLM API, including temperature and sampling parameters. While this reflects common usage patterns and hence enhances ecological validity, these parameters are known to influence output variability and originality \citep{peeperkorn_is_2024}. Future research should systematically vary such parameters to explore their effects on divergent thinking performance.

Second, we focused exclusively on English-language tasks and did not investigate multilingual capabilities. Given the global reach of GenAI tools and recent evidence suggesting that LLMs may behave differently across languages \citep{zhang_dont_2023}, this remains an important area for future exploration. Third, while we evaluated a wide range of models on standardized DT tasks, we did not address more complex aspects of creativity such as problem finding, iterative elaboration, or social validation—dimensions often considered central in high-level creative work \citep{amabile_pursuit_2017, simonton_what_2013}. As such, our findings should be interpreted as reflecting creative potential rather than holistic creative capacity.

Ethical considerations also arise from our findings. On the one hand, the ability of LLMs to generate original content that rivals or exceeds average human output suggests a democratizing potential: these tools can help individuals with limited experience engage in creative expression. On the other hand, the proliferation of AI-generated content raises concerns about authenticity, plagiarism, and the dilution of originality in digital culture. Repeated exposure to AI-generated ideas may subtly influence human ideation, narrowing the perceived range of what is creative or acceptable --potentially reinforcing the very homogenization effects observed in this study and elsewhere \citep{toma_effects_2024, doshi_generative_2024}. Moreover, while exposure to others’ ideas has been shown to benefit human creativity \citep{fink_creative_2009}, the role of GenAI in this dynamic is more complex. Unlike human peers, LLMs generate content that is simultaneously vast in volume and narrow in conceptual distribution. This tension necessitates thoughtful integration of GenAI into educational, artistic, and professional workflows, ideally with mechanisms that encourage critical thinking and safeguard human agency.

As GenAI continues to reshape creative practice, the importance of maintaining meaningful human oversight cannot be overstated. Responsible use requires not only technological literacy but also ethical awareness and reflective practices that ensure AI augments rather than undermines human creative expression.

\section{Acknowledgments}

We would like to thank the \href{https://www.zib.de}{Zuse Institute Berlin} for hosting various LLM models for testing and Peter Organisciak for providing us with an API key for the OpenScoring API. We would also like to thank the authors of \citet{olson_naming_2021} and \citet{organisciak_beyond_2023} for making their work and codes, including their scoring sites, publicly available. 

Research reported in this paper was partially supported by the Deutsche Forschungsgemeinschaft (DFG) through the DFG Cluster of Excellence MATH+ (grant number EXC-2046/1, project ID 390685689), and by the German Federal Ministry of Education and Research (BMBF), grant number 16DII133 (Weizenbaum-Institute).

\bibliographystyle{apalike} 
\bibliography{references, extras}

\begin{landscape}
\section{Appendix}

\renewcommand{\arraystretch}{1.3} 
\setlength{\tabcolsep}{6pt} 
\begin{longtable}{>{\raggedright\arraybackslash}p{2.5cm} 
                  >{\raggedright\arraybackslash}p{2.3cm} 
                  >{\raggedright\arraybackslash}p{2.8cm} 
                  >{\raggedright\arraybackslash}p{3cm} 
                  >{\raggedright\arraybackslash}p{1cm} 
                  >{\raggedright\arraybackslash}p{4.5cm} 
                  >{\raggedright\arraybackslash}p{5.7cm}}
\caption{Overview of Studies Measuring the Creativity of Large Language Models.}\label{tab:creativity_overview}\\
\\
\hline\hline
\textbf{Paper (First Author, Year)} & \textbf{Creativity Measure Used} & \textbf{AI Model Used} & \textbf{Method to Assess Creativity} & \textbf{Comp. Humans} & \textbf{Results} & \textbf{Key Findings} \\
\hline
\endfirsthead

    \multicolumn{7}{l}{\textbf{LLMs Outperform Humans}} \\
    \midrule
    \cite{bellemare-pepin_divergent_2024} & DAT, Divergent Semantic Integration (DSI) & StableLM, Pythia, RedPajama, GPT-4, GPT-4-turbo, GPT-3, Vicuna, Claude3, GeminiPro & Semantic creativity comparison between LLMs and 100,000 humans & $\nearrow$ & GPT-4 outperformed humans on DAT, but humans had higher DSI in creative writing & LLMs generated more semantically distant associations than humans but lacked deep contextual integration. Humans produced richer, more meaningful associations in storytelling and open-ended tasks. Prompting makes a difference. \\
    \cite{castelo_how_2024} & Product Design Creativity & GPT-4 & PCA of novelty, originality, and usefulness & $\uparrow$ & GPT-4 ideas rated as more creative than human-generated ones across multiple experiments & AI-generated product ideas outperformed human ones in perceived creativity, showing stronger effects in originality. \\
    \cite{cropley_is_2023} & DAT & ChatGPT-3.5, ChatGPT-4 & AI-generated responses scored using semantic distance model & $\nearrow$ & GPT4, has, on average, a higher verbal divergent production capability than most (85\% of) humans. GPT4, has, on average, a higher verbal divergent production capability than most (85\% of) humans & GPT-4 outperformed humans but was unreliable and repetitive. \\
    \cite{girotra_ideas_2023} & Consumer Product Idea Generation & GPT-4 & Comparison of AI- and human-generated product ideas using consumer purchase intent surveys & $\uparrow$ & AI-generated ideas had higher average purchase intent than human-generated ideas & AI-generated ideas were more numerous and rated higher in quality and feasibility. AI exhibited greater variance in idea quality. \\
   \cite{gomez-rodriguez_confederacy_2023} & Human Evaluation of Creative Writing & GPT-4, Claude, ChatGPT-3.5, bing, koala, vicuna, oa, bard, stableLM, dolly, alpaca & Expert grading of LLM-generated vs. human-authored stories using a 10-item rubric & $\nearrow$ & GPT-4 outperformed human writers in readability and structure, while humans excelled in originality & AI-generated texts were coherent and grammatically refined but lacked thematic depth and genuine emotional engagement. Humans maintained an advantage in originality and narrative unpredictability. \\
   \cite{guzik_originality_2023} & TTCT & ChatGPT-4 & Fluency, flexibility, and originality scores compared to human norms & $\uparrow$ & AI in top 1\% for originality and fluency & AI outperformed humans in fluency, flexibility, and originality. \\
   \cite{haase_artificial_2023} & AUT & ChatGPT-3, ChatGPT-4, alpa.ai, Copy.ai, Studio.ai, YouChat & AI and human-rated originality & $\nearrow$ & no qualitative difference between AI and humangenerated creativity; 9.4\% of humans were more creative than the most creative GAI & AI matched human-level creativity in divergent thinking tasks. \\
   \cite{hubert_artificial_2023} & AUT, CT, DAT & GPT-4 & AI vs. human performance on multiple divergent thinking tasks & $\uparrow$ & $\eta^2 = .51$ (AUT), .49 (CT), .16 (DAT) & AI outperformed humans in originality and elaboration. \\
   \cite{koivisto_best_2023} & AUT & ChatGPT-3.5, ChatGPT-4, Copy.ai & Semantic distance and human-rated originality & $\uparrow$ & AI scored higher on average, but best human responses exceeded AI & AI responses were high-quality and varied, but humans showed more unpredictability and conceptual leaps, outperforming AI in originality and depth. \\
   \cite{summers-stay_brainstorm_2023} & AUT & GPT-3 & Brainstorming phase followed by a selection phase to enhance response quality & $\uparrow$ & When guided through a two-step process, GPT-3 exceeded human level on AUT & Structured generation and selection significantly enhance AI-generated creative responses. \\
    \newpage
    \midrule
    \multicolumn{7}{l}{\textbf{LLMs are similar to Human performance}} \\
    \midrule
   \cite{gilhooly_ai_2023} & AUT & GPT-3 & Human judges rated originality, usefulness, and surprise & $\rightarrow$ & Small differences & Humans rated slightly higher on originality and flexibility; GPT-3 performed better on usefulness. \\
   \cite{nath_characterising_2024} & AUT, Verbal Fluency Task & GPT-4 Turbo, Claude-3, Gemini-1.0 Pro, Palm, Meta 70B Llama 3, Mistral AI 7B, NousResearch 7B, Upstage 10.7B SOLAR & Automated analysis of response pathways (persistent vs. flexible) using sentence embeddings and semantic similarity metrics  & $\rightarrow$  &  AI responses matched human response distributions on AUT but showed biases toward either persistence or flexibility. In VFT, LLMs were significantly more persistent than humans.  & AI demonstrated variability in semantic search strategies, with some models favoring deep search in constrained spaces (persistent) and others favoring broad jumps across categories (flexible). In humans, both pathways led to comparable creativity scores, whereas for AI, flexibility correlated more strongly with originality. \\
  \cite{orwig_does_2024} & Short Story Task, Semantic Distance & GPT-3, GPT-4 & Creativity ratings based on semantic diversity and perceptual details & $\rightarrow$ & r = .56 (semantic diversity), r = .16 (perceptual details) & Human and AI-generated stories were rated similarly in creativity \\
   \cite{si_can_2024} & Research Ideation Evaluation & Claude-3.5, GPT-4 & Blind-reviewed comparison of human and AI-generated novel research ideas & $\rightarrow$ & AI-generated ideas rated as significantly more novel than expert ideas but slightly weaker on feasibility & AI-generated research ideas were found to be more novel than those written by human NLP experts. However, feasibility remains a challenge, and AI lacks diversity in idea generation. \\
  \cite{sun_large_2024} & Multi-Domain Creative Task Evaluation & GPT-3.5, GPT-4, Claude, Qwen, SparkDesk & Benchmarking LLMs against humans in divergent thinking, problem-solving, and creative writing & $\rightarrow$ & LLMs ranked in the 52nd percentile against humans, excelling in problem-solving but underperforming in creative writing & When asked 10 times, one LLM equaled 8-10 humans in collective creativity. LLMs performed well in divergent thinking but lacked diversity and feasibility in novel idea generation. \\
   \cite{zhao_assessing_2024} & TTCT & GPT-4, LLaMA-2-70B, Qwen, Vicuna & 7-task benchmark evaluating Fluency, Flexibility, Originality, and Elaboration in AI-generated responses & $\rightarrow$ & LLMs excel in elaboration but perform weakest in originality; multi-agent collaboration improves creativity & AI-generated responses are elaborative but struggle with original thought. Multi-agent collaboration slightly boosts originality. Creativity varies significantly between different LLM architectures and training paradigms. \\
     \midrule
     \multicolumn{7}{l}{\textbf{LLMs perform worse than Humans}} \\
    \midrule
   \cite{chakrabarty_art_2023} & TTCW & GPT-4, Claude, ChatGPT & Expert evaluations on short fiction writing & $\downarrow$ & LLMs fail TTCW tests 3-10x more often than humans & GPT-4 performs best in originality but lacks coherence and elaboration. \\
   \cite{grassini_artificial_2024} & Figural Interpretation Quest & ChatGPT-4 & Semantic distance and human ratings of subjective creativity & $\downarrow$ & AI more flexible, but humans scored higher on perceived creativity & AI responses were flexible but perceived as less novel and meaningful than human responses, which excelled in context-aware interpretation. \\
    \cite{ismayilzada_evaluating_2024} & Creative Short Story Task & GPT-4, Gemini-1.5, Claude-3.5, Llama-3.1-405B & Creativity evaluation using novelty, surprise, and diversity metrics & $\downarrow$ & Humans significantly outperformed AI in all metrics & AI-generated stories were stylistically complex but lacked novelty, surprise, and diversity. \\
    \cite{lu_benchmarking_2025} & NEOGAUGE Benchmarking (Divergent \& Convergent Thinking) & GPT-4, Claude, LLaMA & Evaluation via DENIAL PROMPTING and NEOGAUGE metric & $\downarrow$ & GPT-4 most creative, but below human-level & AI exhibited limited problem-solving creativity \\
    \cite{marco_small_2025}& Creative Fiction Writing Evaluation & BART-large, GPT-3.5, GPT-4o & Human study comparing AI vs. human-written short stories based on grammaticality, relevance, creativity, and attractiveness & $\downarrow$ & SLM outperformed humans in most dimensions except creativity (not statistically significant) & BART-large, a fine-tuned small language model (SLM), outperformed human writers in all assessed aspects except creativity, where humans held a slight, non-significant advantage. Smaller models can rival both humans and larger models in creative writing under certain conditions. \\
   \cite{stevenson_putting_2022}& AUT & GPT-3 & Expert-rated originality, usefulness, and surprise; semantic distance scoring & $\searrow$ & Humans had higher originality, surprise, and flexibility scores; GPT-3 had higher utility ratings & GPT-3 performed well but was outperformed by humans in originality and flexibility. Utility scores were higher for GPT-3. Flexibility in response patterns varied significantly. \\
   \cite{vinchon_artificial_2023}& Evaluation of Potential Creativity (EPoC battery) & ChatGPT-3.5, ChatGPT-4 & Standardized narrative creativity tasks, CAT evaluation & $\downarrow$ & AI performed well in fluency, but lacked originality & GPT-4 showed high fluency but lacked originality in story generation. AI-generated stories were rated as lower in diversity and novelty compared to human creative works.  \\
   \cite{wenger_were_2025} & AUT, DAT, Forward Flow & AI21 Jamba 1.5 Large, Google Gemini 1.5, Cohere Command R Plus, Meta Llama 3 70B Instruct, Mistral Large, GPT 4o, and Phi 3 medium 128k Instruct & Measuring cross-LLM homogeneity in creative responses using standardized creativity tests & $\downarrow$ & LLM responses significantly more similar to each other than human responses, even after controlling for response structure & Cross-LLM homogeneity suggests that all LLMs converge towards a limited range of creative outputs, limiting divergent thinking regardless of model choice. Increased creativity prompts slightly improved response diversity but not to human levels. \\
    
     \bottomrule
\end{longtable}
\textbf{Notes:}  
AUT - Alternative Uses Task;
CT - Consequences Task; 
DAT - Divergent Association Task; 
TTCT - Torrance Tests of Creative Thinking; 
TTCW - Torrance Test for Creative Writing 
\end{landscape}
\newpage

\subsection{Divergent Association Task - additional plots}

\newpage
\subsection{Alternate Use Task - additional plots}

\end{document}